\documentclass{article} 
\usepackage[final]{neurips_2026}
\usepackage{times}

\makeatletter
\@ifundefined{@noticestring}{}{\renewcommand{\@noticestring}{}}
\@ifundefined{@notice}{}{\renewcommand{\@notice}{}}
\makeatother

\usepackage{amsmath,amsfonts,bm}

\def\eqref#1{equation~\ref{#1}}

\def\1{\bm{1}}

\DeclareMathAlphabet{\mathsfit}{\encodingdefault}{\sfdefault}{m}{sl}
\SetMathAlphabet{\mathsfit}{bold}{\encodingdefault}{\sfdefault}{bx}{n}

\usepackage{hyperref}
\usepackage{url}
\usepackage{graphicx}
\usepackage{wrapfig}
\usepackage{natbib}
\usepackage{xcolor}
\usepackage{tcolorbox}
\usepackage{booktabs}
\usepackage{multirow}
\usepackage{bm}
\usepackage[textsize=footnotesize,textwidth=1.1in,shadow,loadshadowlibrary]{todonotes}
\usepackage[normalem]{ulem}   
\usepackage{array}
\usepackage{capt-of}
\usepackage{fontawesome5}  

\usepackage{amsmath}
\usepackage{amsfonts}

\definecolor{prasanncolor}{RGB}{192,57,43}     
\definecolor{amandacolor}{RGB}{39,110,170}     
\definecolor{jacobcolor}{RGB}{39,140,65}       
\definecolor{sewoncolor}{RGB}{142,68,173}      

\definecolor{claudeeditcolor}{RGB}{230,126,34}  

\newcommand{\ctc}[1]{ \textcolor{amandacolor}{$O_{T}(#1)$}}

\definecolor{autocolor}{RGB}{200,110,20}        

\definecolor{claudecolor}{RGB}{124,74,32}       
\newcommand{\claude}[1]{\textcolor{claudecolor}{#1}}

\title{No More Free Lunch: \\Corpus Task Complexity Matters as Corpora Grow}

\definecolor{affblue}{HTML}{2E4BDE}   
\definecolor{affpink}{HTML}{E0218A}   
\definecolor{affred}{HTML}{B01C32}    
\definecolor{affviolet}{HTML}{7A4FD6} 

\newcommand{\amark}{\textcolor{affblue}{\textsuperscript{$\alpha$}}}
\newcommand{\bmark}{\textcolor{affpink}{\textsuperscript{$\beta$}}}
\newcommand{\gmark}{\textcolor{affred}{\textsuperscript{$\gamma$}}}

\newcommand{\amarki}{\textcolor{affblue}{$^{\alpha}$}}
\newcommand{\bmarki}{\textcolor{affpink}{$^{\beta}$}}
\newcommand{\gmarki}{\textcolor{affred}{$^{\gamma}$}}

\newcommand{\resourcelinks}{%
\begin{center}
\begin{tabular}{@{}c@{\hspace{0.9em}}l@{\hspace{1.6em}}l@{}}

\faGithub  & \textbf{Code / Data}          & \texttt{\href{https://github.com/PrasannS/corpustaskcomplexity}{PrasannS/corpustaskcomplexity}} 
\end{tabular}
\end{center}}

\author{%
  Prasann Singhal\amark\quad
  Amanda Bertsch\gmark\bmark\quad
  Jacob Steinhardt\amark \quad
  Sewon Min\amark\bmark
  \\[0.45em]
  \normalfont\small
  \amarki UC Berkeley \quad
  \bmarki Allen Institute for AI \quad
  \gmarki Carnegie Mellon University 
  \\[0.30em]
  \normalfont\small\ttfamily
  prasann@berkeley.edu
}

\usepackage{xspace}
\newcommand{\cname}{CTC\xspace}
\newcommand{\bench}{\textsc{CTC-Bench}}

\begin{document}

\maketitle

\resourcelinks

\begin{abstract}

Given a large corpus, the questions one might ask can vary—from ``When was the first human heart transplant?'' to ``What are all the contradictory claims in this literature?''---but what makes some questions more challenging than others? In this work, we define a notion of \textbf{Corpus Task Complexity (CTC)} that characterizes tasks by how their difficulty grows with corpus size; for instance, a retrieval query only requires a single linear pass over a corpus, while finding contradictions requires checking a quadratically growing set of claim pairs. Observing that prior work has largely only studied tasks whose difficulty grows linearly with corpus size, which we call \emph{low} CTC tasks, we introduce 10 new tasks belonging to a class of \emph{high} CTC whose difficulty grows quadratically or more in corpus size. We find that high-CTC tasks not only grow much more challenging on average at longer contexts for LCLMs, they reverse many modeling conclusions drawn solely from low-CTC evaluations. For instance, efficient block-sparse and hybrid attention approaches consistently match full attention performance on low-CTC tasks, but degrade much more on high-CTC tasks. Large-corpus high-CTC reasoning thus remains an open challenge as full attention is too costly to scale, motivating future research on these tasks. We release our code, data, and 22-task suite (\bench), to facilitate future research in this area.

\end{abstract}

\section{Introduction}


Computational tools enable us to conduct search over large digital corpora---from scientific literature to the internet---but is it yet possible to develop systems that could, for example, \emph{find all contradictions} in a large corpus? Given the wealth of valuable information contained in large corpora, researchers have built diverse systems to extract and analyze this information, from work in information retrieval \citep{TREC2021, Hou2025CLERCAD} to summarization \citep{Kocisk2017TheNR}. We refer to these collectively as \emph{corpus reasoning} tasks, but they can vary substantially in their complexity over the corpus, ranging from simple factoid queries like ``When was Ralph Lauren founded?'', to tasks requiring extensive cross-document interaction, such as ``Find all contradicting claims in this biology literature'' or ``What are the outliers in the LLM agent trace corpus?''.  

Long-context language models (LCLMs) trained to process large inputs (e.g., 32K tokens or more) are promising for corpus tasks---with softmax attention enabling direct retrieval and reasoning across parts of the conditioned corpus. Modern LCLMs achieve strong performance on many existing synthetic long-context benchmarks~\citep{Yen2024HELMETHT, Bai2024LongBenchVT, Chen2026LongBenchPA}, even with various efficient approaches used to reduce their computational cost~\citep{Yang2024GatedDN, Beltagy2020LongformerTL}. But given the lack of clear characterization of what makes corpus reasoning tasks difficult on large corpora, it remains unclear whether these successes apply to all corpus reasoning tasks, particularly more complex tasks.

\begin{figure}[t]
    \centering
    \begin{minipage}[b]{0.95\textwidth}
        \includegraphics[width=\textwidth, ]{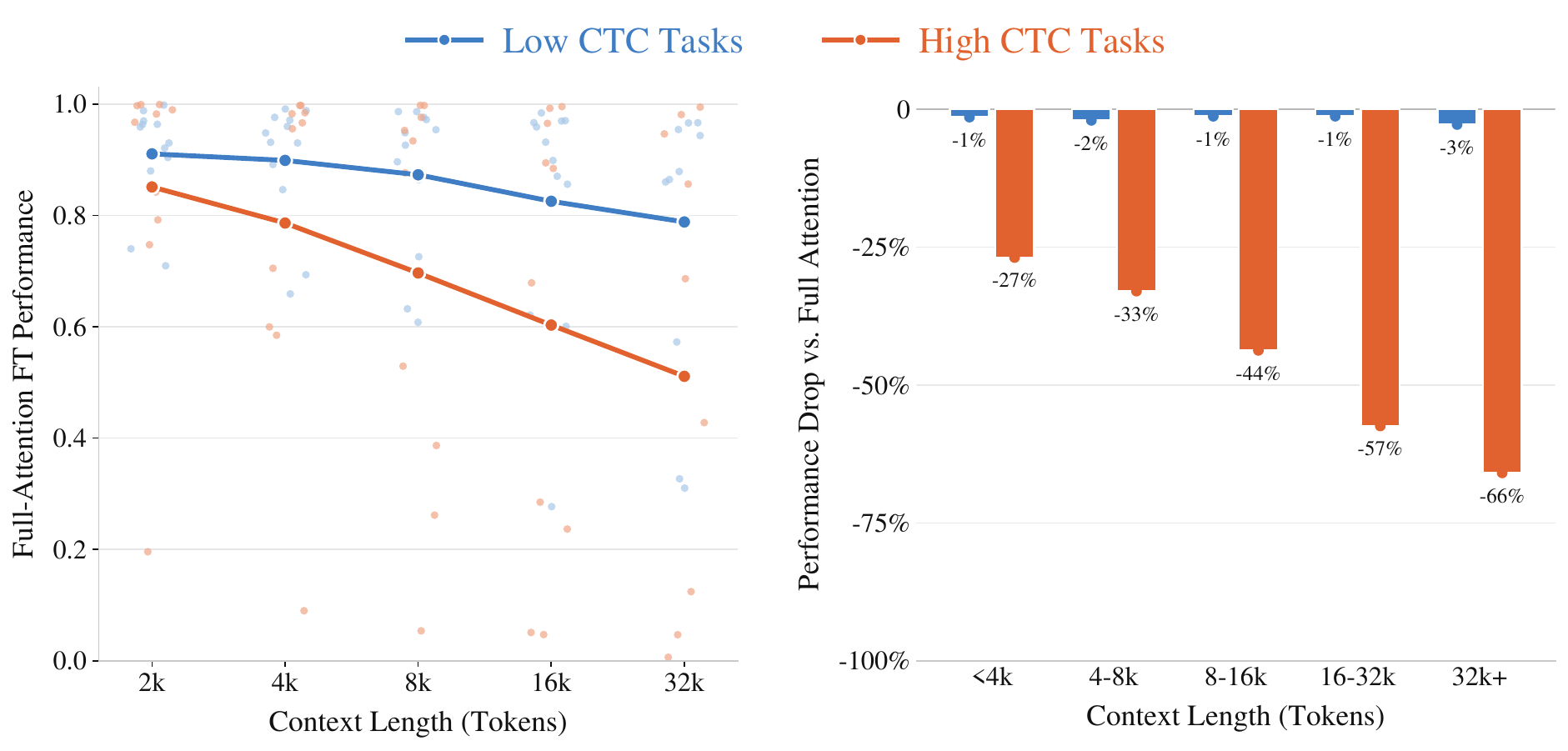}\\
    \end{minipage}
    \caption{
    Performance of in-distribution trained Qwen3.5-4B models on low- and high-\cname\ tasks.
    \emph{(Left)} Performance degrades faster on high-\cname\ tasks than on low-\cname\ tasks as corpus size increases.
    \emph{(Right)} Block-sparse attention causes almost no performance degradation relative to full attention on low-\cname\ tasks, but substantial degradation on high-\cname\ tasks.
}

    \label{fig:motivatingtask}
\end{figure}



We begin by defining \emph{corpus task complexity} (CTC), which characterizes a task by how the number of operations required to solve it scales with corpus size. For example, the number of local operations required to retrieve a fact usually grows linearly with corpus size, since each document need only be checked once to find where the fact is located. In contrast, a task such as ``Find all contradicting claims in the literature'' involves comparing all pairs of claims, resulting in quadratic growth. Through this lens, we survey 12 diverse corpus reasoning tasks commonly used to evaluate LCLMs and find them to all grow linearly in difficulty with corpus size (which we call \emph{low-CTC}). We then introduce 10 diverse tasks whose complexity grows quadratically or faster (\emph{high-CTC} tasks), ranging from identifying all contradictory claims in a corpus to finding documents with the rarest topics.

Using this benchmark suite, \bench, we evaluate a range of LCLMs across corpus sizes from 2K to 32K tokens. We first find that with increasing corpus size, performance degrades faster on average on high-CTC tasks than on low-CTC tasks (Figure~\ref{fig:motivatingtask}(a)). More importantly, however, \textbf{we find that high-CTC task results frequently reverse modeling conclusions} drawn solely from low-CTC evaluations. For example, block-sparse attention---an efficient attention variant often used for corpus reasoning~\citep{Xiao2025EfficientLCICL}---shows minimal degradation on low-CTC tasks, yet on high-CTC tasks it sees large performance drops relative to full attention (Figure~\ref{fig:motivatingtask}(b)). Similar discrepancies occur for other modeling decisions: hybrid architectures that alternate linear and full attention exhibit larger performance gaps on high-CTC tasks; and length generalization---fine-tuning a long-context model on shorter contexts than the intended context length at test time~\citep{Gao2024HowTT}---can work for low-CTC tasks but breaks down on high-CTC tasks. 

To summarize, our results show that evaluations including high-CTC tasks reveal hidden costs to many common modeling choices, and that there is no free lunch: full attention becomes quadratically intractable on large corpora, but efficient attention mechanisms that appear lossless under existing low-CTC evaluations can lead to substantial degradation on high-CTC tasks. Solving high-CTC tasks on large corpora is a challenging long-term problem, and we encourage more research into scalable architectures that can handle both low \emph{and} high-CTC tasks. To support this effort, we release \bench, along with our code and data.

\section{Corpus Task Complexity \& \bench}

\begin{figure}[t]
    \centering
    \includegraphics[width=\textwidth, trim=50 20 50 20, clip]{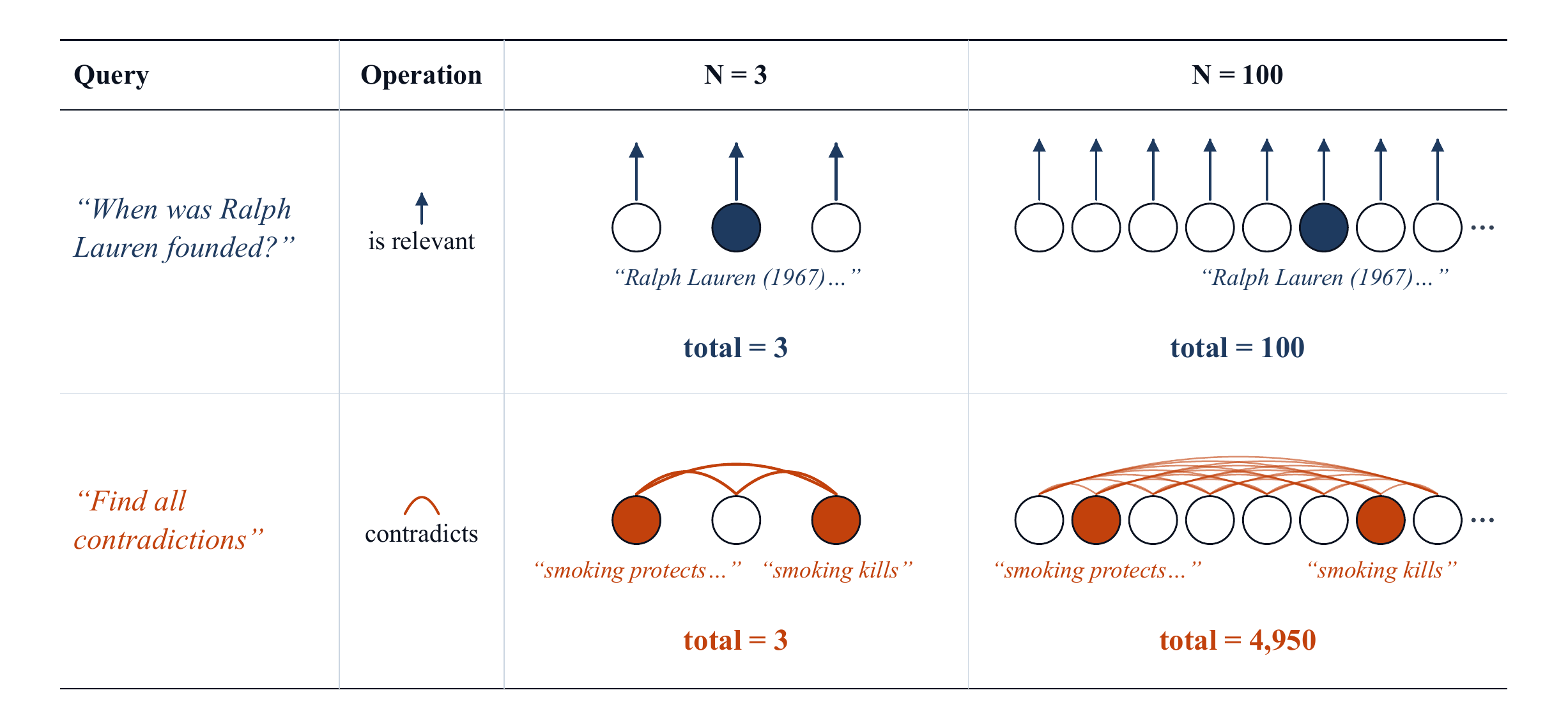}\\
    \caption{Circles represent documents, with shaded circles for the gold documents. As corpus size $N=|C|$ grows from 3 documents to 100 documents, a contradiction search task may in worst-case require many more oracle operations (4950) than factoid retrieval (100).}
    \label{fig:ctcexplanation}
\end{figure}



What most strongly determines corpus task difficulty? We propose \emph{Corpus Task Complexity (CTC)}, a definition of task difficulty based on how a task's difficulty scales with corpus size (\S\ref{sec:definitioncomp}), which we walk through with two tasks (shown in Figure~\ref{fig:ctcexplanation}). We find that most existing benchmarks are low CTC---their complexity grows linearly with corpus size (\S\ref{sec:existingtasks})---then introduce a new class of high \cname tasks, whose complexity grows quadratically or faster (\S\ref{sec:newtasks}).

\subsection{Corpus Task Complexity}
\label{sec:definitioncomp}

Define a corpus $C$ to be an unstructured set of $N$ documents $\{d_1, \ldots, d_N\}$, where each $d_i$ is a sequence of tokens. A corpus reasoning task $T$ is a set of same-type corpus-question-answer samples $(C_i, q_i, a_i) \in T$. For instance a factoid retrieval task sample could be ($C_i=$Wikipedia, $q_i=$\emph{``When was Ralph Lauren founded?''}, $a_i$=\emph{``1967''}), and a contradiction task sample could be ($C_i=$CommonCrawl, $q_i=$ \emph{``Find all contradictions''}, $a_i=$list of contradicting pairs within $C_i$). 

To understanding corpus reasoning tasks, we define a notion of complexity called \emph{corpus task complexity (CTC)}. Imagine that we can call an \textit{oracle} (e.g. an LM-judge) that can answer questions over small contexts with O(1)-length in corpus size. CTC is the asymptotic growth in oracle calls needed to solve tasks as a function of corpus size N.

Specifically, let $A_T$ be an algorithm composed of oracle calls that solves \textit{all} $(C_i, q_i, a_i) \in T$, where $f_{A_T}(C_i, q_i)$ indicates oracle call count, and we constrain $A_T$ to be the algorithm with lowest possible call count across samples. Task $T$ is \ctc{g(N)} if there exists some constant $\alpha$ such that $f_{A_T}(C_i, q_i) \leq \alpha g(N)$ for all $(C_i, q_i, a_i) \in T$. In practice, as proving the optimal $A_T$ can be challenging, we classify CTC with \emph{candidate} best algorithms for tasks given unstructured corpora.


\textbf{Example 1: } (Figure~\ref{fig:ctcexplanation}, blue) To solve factoid retrieval queries like $q=$ \emph{``When was Ralph Lauren founded?''} with oracle calls, $A_T$ may be to run $\mathrm{oracle}(\mathrm{isrelevant},  d_i, q)$ individually on all $d_i\in C$. Since $f_{A_T}(C_i, q_i)$ is just $N$, and less operations aren't possible, this task is \ctc{N}. \footnote{CTC should \emph{not} be confused with individual online or offline cost, which combine to become overall cost. For example, nearest neighbors search often has $O(log(N))$ \emph{query-time} complexity, but still requires $O(Nlog(N))$ \emph{build-time} cost.}

\textbf{Example 2:} (Figure~\ref{fig:ctcexplanation}, orange) To \emph{``Find all contradicting claim pairs.''} with oracle calls, $A_T$ may be to run $\mathrm{oracle}(\mathrm{contradicts}, d_i, d_j)$ on all viable document pairs. While priors or corpus structure could lower this, exhaustively checking \emph{all} possible pairs will generally require quadratic oracle calls in corpus size (e.g. 4,950 calls for 100 documents)---making the task \ctc{N^2}.

Since CTC corresponds to minimal computation \emph{necessary} to solve tasks with growing $N$, we posit that it will correspond to empirical trends for even black-box methods applied to corpus tasks (e.g. LCLMs, dense retrievers). Similar methods may work for tasks within the same CTC, and on larger corpora an \ctc{N^2} task may demand methods with greater inference computation. 

Not all same-CTC tasks are identical---retrieval queries with oracle calls involving exact word matches are harder in a constant sense than ones involving abstract relationships, and such differences may even multiply at higher CTC. But by defining task CTC we can better explain what underlies a task's difficulty, and make better modeling decisions to solve it.

\subsection{Existing Tasks Through the Lens of \cname}

\label{sec:existingtasks}

\providecommand{\ctc}[1]{\ensuremath{\mathcal{O}_T(#1)}}
\providecommand{\cname}{CTC}

\providecommand{\ctc}[1]{\ensuremath{\mathcal{O}_T(#1)}}
\providecommand{\cname}{CTC}
\ifdefined\catw\else\newlength{\catw}\newlength{\dsw}\newlength{\metw}\newlength{\descw}\fi

\providecommand{\ctc}[1]{\ensuremath{\mathcal{O}_T(#1)}}
\providecommand{\cname}{CTC}
\ifdefined\catw\else\newlength{\catw}\newlength{\dsw}\newlength{\metw}\newlength{\descw}\fi

\providecommand{\ctc}[1]{\ensuremath{\mathcal{O}_T(#1)}}
\providecommand{\cname}{CTC}
\ifdefined\catw\else\newlength{\catw}\newlength{\dsw}\newlength{\metw}\newlength{\descw}\fi

\newcommand{\panelcols}{%
  \setlength{\catw}{\dimexpr0.27\linewidth-2\tabcolsep\relax}%
  \setlength{\dsw}{\dimexpr0.33\linewidth-2\tabcolsep\relax}%
  \setlength{\descw}{\dimexpr0.40\linewidth-2\tabcolsep\relax}%
}
\newcommand{\blk}[1]{\multicolumn{3}{@{}l}{#1} \\}
\providecommand{\ex}[2]{#1~$\rightarrow$~\texttt{#2}}

\begin{table}[t]
\centering
\scriptsize
\setlength{\tabcolsep}{4pt}
\renewcommand{\arraystretch}{1.12}

\begin{tabular}{@{}
  >{\raggedright\arraybackslash}p{0.18\linewidth}
  >{\raggedright\arraybackslash}p{0.29\linewidth}
  >{\raggedright\arraybackslash}p{0.18\linewidth}
  >{\raggedright\arraybackslash}p{0.29\linewidth}
@{}}
\toprule
\multicolumn{2}{c}{\textbf{\(\mathbf{O}(N)\): Low Complexity Tasks}} &
\multicolumn{2}{c}{\textbf{Higher-complexity tasks}} \\
\cmidrule(r){1-2}\cmidrule(l){3-4}
\textbf{Dataset} & \textbf{Description} &
\textbf{Dataset} & \textbf{Description} \\
\midrule

NIAH-contra
  & Find the contradicting claim
  & Contradiction
  & Find all contradicting claim pairs \\

SciFact
  & Retrieve evidence for a scientific claim
  & X-Absence
  & Find unmatched docs across 2 shuffled, near-identical corpora \\

FiQA
  & Retrieve relevant financial opinions
  & QDmatch (HPQA)
  & Match questions to two documents \\

MS MARCO
  & Retrieve relevant web passages
  & QDmatch (NQ)
  & Match questions to documents \\

OBLIQ
  & Retrieve passages for subjective queries
  & QDmatch (FiQA)
  & Match financial queries to documents \\

NQ
  & Retrieve the answer passage
  & Outlier (scale $M$)
  & Find chunks from the rarest article \\

HotpotQA 
  & Retrieve two supporting passages
  & Grouping (OpenAlex)
  & Group abstracts by a latent topic \\

MS MARCO rerank
  & Re-rank passages by relevance
  & Strmatch
  & Find pairs sharing a word sequence \\

OOLONG
  & Aggregate labels over the corpus
  & Reorder
  & Recover the order of shuffled segments \\

Outlier (Amazon)
  & Find reviews with the minority attribute
  & Textgroups
  & Find passage triples satisfying a target \\

Outlier (fixed \(M\))
  & Find items from the rarest article
  & & \\

Absence
  & Identify content missing from a copy
  & & \\

\bottomrule
\end{tabular}

\caption{\textbf{Overview of the 22 \bench{} Tasks.}
Linear-time tasks are shown on the left; tasks requiring higher-order comparisons are shown on the right (details, examples in Appendix~\ref{app:examples})
}
\label{tab:eval-datasets}
\end{table}

Which \cname categories do existing corpus reasoning tasks belong to? To answer this, we examine a range of 12 well-established existing tasks. We introduce tasks in the order shown in Table~\ref{tab:eval-datasets} (left):


Information retrieval is arguably the best-studied corpus reasoning task. Within it we examine a synthetic recall \emph{NIAH-contra} task, standard \emph{SciFact, FiQA, and MS-MARCO} tasks taken from BEIR \citep{Thakur2021BEIRAH}, and the twitter set of the modern \emph{OBLIQ} \citep{Tchuindjo2026OBLIQBenchEO} benchmark. While ranging in difficulty, as discussed in our earlier factoid example these tasks are \ctc{N}.

We include retrieval variants of two factoid QA datasets: NaturalQuestions (\emph{NQ}) \citep{Kwiatkowski2019NaturalQA}, a single-hop task, and HotpotQA \citep{Yang2018HotpotQAAD}, a multi-hop task. Both call $\mathrm{oracle}(\mathrm{hasanswer}, q, d_i)$. Since, even in the multi-hop case requiring multiple retrieval steps, only a constant number of passes are required---one retrieval step to find the first hop, and one retrieval step to find the second hop---we classify these also as \ctc{N}. Likewise, MSMARCO rerank, which involves finding top-10 most relevant documents to a query, given a small k, is at most a constant number of retrieval passes, and thus is also \ctc{N}.


We include several corpus aggregation tasks. \textit{OOLONG}\citep{Bertsch2025OolongEL} involves distributional or counting questions about class labels of documents. \textit{Outlier (Amazon)} involves finding the least common star or product category in a set of reviews. Given a fixed set of class labels, a single pass of $\mathrm{oracle}(\mathrm{classify}, d_i)$ (e.g. 2-stars, 4-stars) calls with lightweight aggregation is typically sufficient for these tasks, making them \ctc{N}. Lastly, we include a 2-corpus \textit{Absence} task based on AbsenceBench \citep{Fu2025AbsenceBenchLM}---searching for deletions in near-identical Project Gutenberg passages. Since positions are aligned, the oracle is $\mathrm{oracle}(\mathrm{matches}, d_i, d_{i-N})$, where $N$ is individual corpus size. Only a single pass over the second corpus is needed here, so this is \emph{also} \ctc{N}.

The above tasks constitute a recent, diverse, and representative sample of the most studied IR and LCLM benchmark tasks, yet \textbf{they all fall into \ctc{N}}.

\subsection{\bench{} and High \cname Tasks}

\label{sec:newtasks}

 Our definition of \cname suggests that a variety of useful tasks fall into higher complexity classes, yet such tasks, while explored at small scale (e.g. event re-ordering \cite{Chambers2014DenseEO}), are understudied for larger corpora, especially in the context of LCLMs. We thus, to enable our analysis, construct a diverse suite of 10 high-\cname tasks (Table~\ref{tab:eval-datasets}, right). We call these---along with the 12 low-\cname tasks---\textbf{\bench{}} \footnote{We often include synthetic / semi-synthetic tasks for controllability and cleaner analysis, though we design all \bench{} tasks to structurally resemble realistic tasks (see Appendix~\ref{app:realism}) } :

(a) \emph{Contradiction} $q_i=$\emph{``Find all the contradicting claim pairs in the corpus.''}, $a_i$ is a list of 3 contradicting claim pair IDs. It uses LLM-generated contradictions to pubmed abstract claims \citep{Jin2019PubMedQAAD}, and is as discussed earlier \ctc{N^2} (Section~\ref{sec:definitioncomp}) .

(b) \emph{X-Absence} $q_i=$\emph{``Given near-identical corpus A and B with shuffled chunks, find chunks that are only in one corpus''}, $a_i$ is a list of chunks only in one of either A or B. All $d_i$ are chunks of a Project Gutenberg passage. $C_A$ and $C_B$ ($N$ chunks each) are shuffled to prevent positional bias (unlike Absence which is ordered), so total $\mathrm{oracle}(\mathrm{matches}, d_i, d_j)$ calls are \ctc{N^2}.


(c-e) \emph{(QDmatch)-FiQA, NQ, HotpotQA} $q_i$=\emph{``Given corpus A with single-sentence questions and B with documents, find (question, document) pairs where documents answer questions''}, $a_i$ is a list of 3 matching pair IDs. We use 3 such settings, using retrieval questions and documents from the FiQA, NQ, and HotpotQA tasks to construct this task. Since $C_A$ and $C_B$ are size $N$ each, cross checking all pairs with $\mathrm{oracle}(\mathrm{isrelevant}, d_i, d_j)$ is \ctc{N^2}.

(f) \emph{Outlier (wiki)} $q_i$=\emph{``Given the corpus with chunks discussing different topics, identify chunks belonging to the least common topic''}, $a_i$ is a list of minority topic chunk IDs. We source chunks ($d_i$) from $M$ Wikipedia pages, where the ``topic'' is the page source (e.g. ``Horses''). Unlike OOLONG, which has fixed classes, page topics are diverse and don't repeat across corpora. Given $\mathrm{oracle}(\mathrm{samecategory}, d_i, d_j)$, since each new document only needs to check one document per category, this is \ctc{NM}. To test this we include an Outlier (fix $M$) setting where $M$ stays constant (\ctc{N}) and Outlier (scale $M$) where $M$ grows proportional to $N$ (\ctc{N^2}).  


(g) \emph{Grouping}, $q_i$=\emph{``Group the following scientific abstracts into $k$ groups''}, $a_i$ is a dictionary which must have $k$ keys and put abstracts in the same topic group together (note $k$ increases naturally with $N$). Abstracts ($d_i$) and true topic labels (e.g. Biology, Plant Biology) are mined from OpenAlex \citep{Priem2022OpenAlexAF}. This has the the same structure as Outlier (wiki) and is \ctc{NM}.

(h) \emph{Strmatch}, $q_i$=\emph{``Find word sequences with at least k words in common''}, $a_i$= 3 pair IDs that match. $d_i$ are random noun sequences. Calling $\mathrm{oracle}(\mathrm{hascommonk}, d_i, d_j)$, the task requires searching over \ctc{N^2} pairs. This task serves as a synthetic control for other pairwise tasks.

(i) \emph{Reorder} (\cite{Bai2024LongBenchVT}), $q_i$=\emph{``Given the passage (chunks have been randomly shuffled), output the original ordering.''}, $a_i$ is an ID list of length $N$ with true ordering, and $d_i$ are sentences from a Project Gutenberg passage. If we use $\mathrm{oracle}(\mathrm{directlyneighbors}, d_i, d_j)$, this requires brute-force checking all pairs to recover the order (\ctc{N^2}).

(j) \emph{Textgroups}, $q_i$=\emph{``Find groups of 3 passages whose adjective count add ups to t.''} given Project Gutenberg chunks, where property and count $t$ vary (e.g. 67 nouns). With calls to $\mathrm{oracle}(\mathrm{propertysum}, d_i, d_j, d_k)$ searching over all triples, we include it as an \ctc{N^3} exemplar.

\section{Role of Task Complexity on Growing Corpora}

\label{sec:experimental_setup}


Having established CTC and \bench{}, we now evaluate a range of models to understand how well they perform on these tasks. In particular, we focus on performance as a function of corpus size: \emph{How challenging do high-\cname tasks become as corpus size grows?}


\textbf{Experimental Setup} We train Qwen3.5-4B individually for all 22 \bench{} tasks, ranging from 2k to 32k context length in training and evaluation. For all tasks except OBLIQ (where we use synthetic data, Appendix~\ref{app:data-details}), we use existing or generated in-domain fine-tuning sets to isolate architectural limits independent of data and generalization.

We always train for 1 epoch on 20k datapoints per task (except for OBLIQ and SciFact), with evenly split examples across context lengths (matching the corpus sizes we evaluate on). We train with full fine-tuning, with a learning rate of $5\times 10^{-5}$. All tasks use this prompt format: 

{\small${\texttt{[Initial]\textcolor{jacobcolor}{\textit{ 1:[Doc 1 content] \ldots N:[Doc N content]}} [Query] \textbf{[Answer]}}}$}

\textbf{Result: High-\cname Task Performance Drops More Rapidly Than Low-\cname.} We plot in Figure~\ref{fig:motivatingtask} (left) \textit{averages across tasks} of full attention performance for all High-\cname (orange) and Low-\cname (blue) tasks across \bench{} (task-individual plots in Figure~\ref{fig:multitask_fig4}). Average performance drops $0.910 \rightarrow 0.788$ (13\%) from 2k to 32k on low-\cname (10 out of 12 tasks degrade less than $0.1$ in performance) but on high-\cname tasks average performance drops $0.851 \rightarrow 0.511$ (40\%) from 2k to 32k (7 of 10 tasks drop over $0.1$ in performance). In aggregate, even with in-distribution training, high complexity task difficulty grows much faster on larger corpora.  



\section{Rethinking Free Lunches With CTC}
\label{sec:methodcomp}




In the previous section, we show that high-\cname tasks often degrade at larger corpus sizes much faster than more commonly studied low-\cname tasks---suggesting that even costly full attention models are insufficient at scale for high-\cname tasks. But how does this relate to the large body of prior LCLM work that has predominantly focused on \emph{reducing} LCLM computation? We revisit 3 common modern LCLM choices from the lens of \cname: block-sparse attention (\S\ref{sec:blocksparse}), hybrid architectures (\S\ref{sec:fullvhybrid}), and length generalization from short train contexts (\S\ref{sec:lengthgen}).  

\subsection{Full vs Block-Sparse Attention}
\label{sec:blocksparse}

\begin{figure}[t]
\centering
\includegraphics[width=\textwidth]{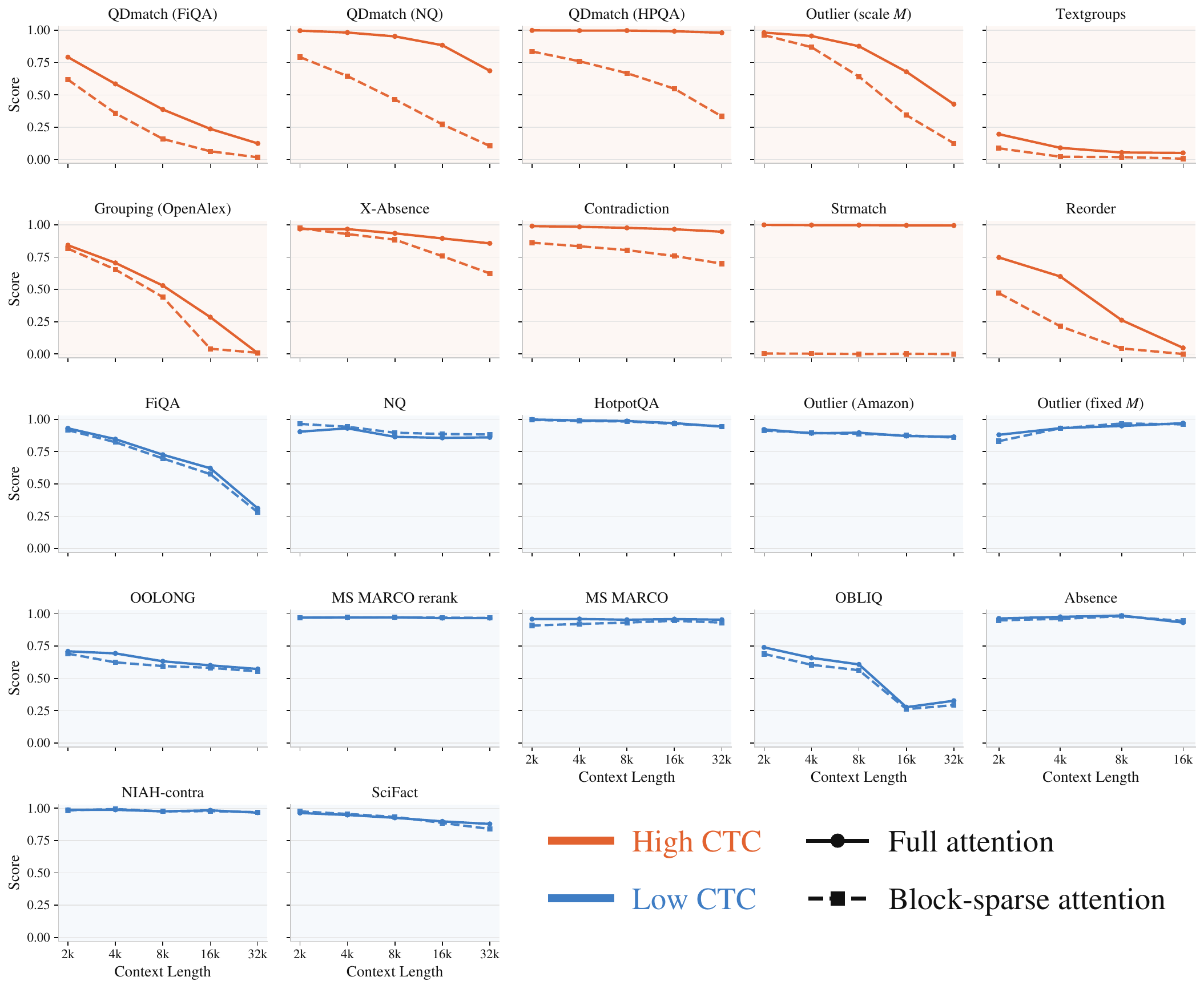}
\caption{\emph{Scaling of full vs.\ block-sparse attention (mask mix)} We show block-sparse vs full attention performance at different contexts on our full task suite. While both methods perform near-identically on low-\cname, high-\cname tasks correspond to growing gaps between them on larger corpora. } 
\label{fig:multitask_fig4}
\end{figure}

In a regular Transformer model (full attention), pre-filling a corpus has quadratic computational complexity. Motivated by the fact that in corpus reasoning, the input is a collection of independent documents, much prior work has tried \citep{Beltagy2020LongformerTL, Zaheer2020BigBT, Gollapudi2026CanLM, Xiao2025EfficientLCICL} block-sparse attention: pre-filling each document independently so that they attend to tokens within the same document only, with only a small number of query and answer tokens attending globally. These approaches have been shown effective, though often limited to evaluation on \ctc{N} settings like QA, retrieval, and in-context learning. 


We investigate several decisions reducing computation from full attention, but what if we instead apply \emph{more} computation to cheap methods? Our block-sparse attention implementation follows that of \citep{Xiao2025EfficientLCICL} \footnote{Qwen3.5 models use 75\% gated delta-net layers (already linear), so we only apply block-sparse to remaining 25\%}, but we additionally propose a new \emph{mask-mixing} technique: during block-sparse attention training, with probability $p$ a training batch uses the full-attention mask instead of the block-sparse one, where $p=0$ indicates original block-sparse attention, and $p=1$ indicates full attention (we anneal on a curriculum from $p=0.8$ to $p=0$ during training). During evaluation, block-sparse attention is used for all inputs. 

\begin{figure}[t]
    \centering

    \begin{minipage}[c]{0.49\textwidth}
        \vspace{0pt}
        \centering
        \small
        \setlength{\tabcolsep}{4pt}
        \renewcommand{\arraystretch}{1.18}

        \resizebox{\linewidth}{!}{%
        \begin{tabular}{lr@{\hspace{1.4em}}lr}
        \toprule
        \multicolumn{2}{c}{\textbf{Low CTC}}
        & \multicolumn{2}{c}{\textbf{High CTC}} \\
        \cmidrule(r{1.2em}){1-2}\cmidrule{3-4}
        Task & Gain & Task & Gain \\
        \cmidrule(r{1.2em}){1-2}\cmidrule{3-4}
        NQ               & $+0.002$ & QDmatch (FiQA)      & $+0.031$ \\
        HotpotQA         & $+0.004$ & QDmatch (NQ)        & $+0.283$ \\
        FiQA             & $+0.011$ & Outlier (scale-$M$) & $+0.505$ \\
        OOLONG           & $+0.581$ & XAbsence            & $+0.646$ \\
        Outlier (Amazon) & $+0.773$ & Contradiction       & $+0.773$ \\
        \cmidrule(r{1.2em}){1-2}\cmidrule{3-4}
        \textit{Mean}    & $+0.274$ & \textit{Mean}       & $+0.448$ \\
        \bottomrule
        \end{tabular}%
        }
    \end{minipage}
    \hfill
    \begin{minipage}[c]{0.44\textwidth}
        \vspace{0pt}
        \centering
        \includegraphics[width=\linewidth]
            {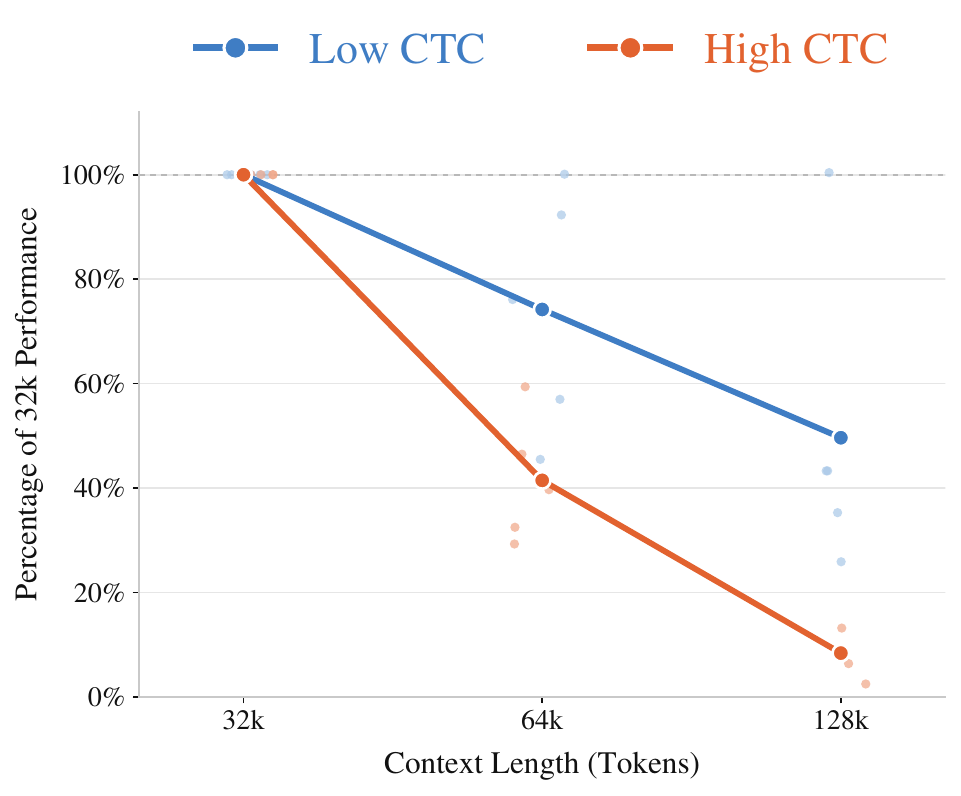}
    \end{minipage}

    \vspace{0.8em}

    \caption{\textbf{(Left)} \emph{Gain of curriculum mask-mixing over pure block-sparse
    training (no mixing)}, averaged across 2k--32k contexts on
    \bench-10. Mask-mixing consistently improves block-sparse, especially
    on High-CTC tasks and tasks more dispersed across contexts (such as
    OOLONG). \textbf{(Right)} \emph{Length generalization (full attention).} For
    64k and 128k context eval sets (trained only to 32k), we plot
    percentage of 32k score. High-CTC tasks are harder for length
    generalization.}
    \label{fig:combined}
\end{figure}

\textbf{Mask-Mixing Frequently Improves Block-Sparse.} We evaluate mask-mixing on a 10 task subset that we call \bench{}-10, that we select to cover diverse task structures and difficulties across CTC classes in \bench{} (results in Figure~\ref{fig:combined}, Left). We find mask-mixing greatly improves block-sparse attention, and \emph{consequently apply it to all block-sparse experiments}. Specifically, while it helps less on retrieval settings like FiQA and NQ, we find that on many High-CTC tasks and tasks requiring aggregation across context like OOLONG, mask-mixing improves block-sparse performance---potentially by distilling representations for tasks from the more computationally expressive full attention.



\textbf{Full vs Block-Sparse Gap Grows Faster in High-CTC.} Figure~\ref{fig:multitask_fig4} shows degradation and comparisons between full and block-sparse attention for the full \bench. On low-\cname tasks, the gap between full and block-sparse attention is negligible \emph{for all tasks}---reproducing much prior work \citep{Beltagy2020LongformerTL, Gollapudi2026CanLM}, we find that block-sparse gives a free lunch of comparable performance (sometimes even better) while being much cheaper. However, across all high-\cname tasks, choosing block-sparse attention 
over full attention leads to worse performance, despite our own improvements through mask-mixing. Unless the task is so difficult that both approach zero (e.g. Textgroups, Reorder), this gap monotonically grows at longer contexts. Averaged across tasks, the performance degradation from using block-sparse on high-\cname{} is $-26.9\%$ (2k), $-33.0\%$ (4k), $-43.7\%$ (8k), $-57.4\%$ (16k) and $-65.9\%$ (32k). This supports a central theme---decisions that did not matter for low-\cname matter greatly on high complexity tasks. 



\label{sec:taskcompresults}

\begin{figure}[t]
\centering
\includegraphics[width=0.95\textwidth]{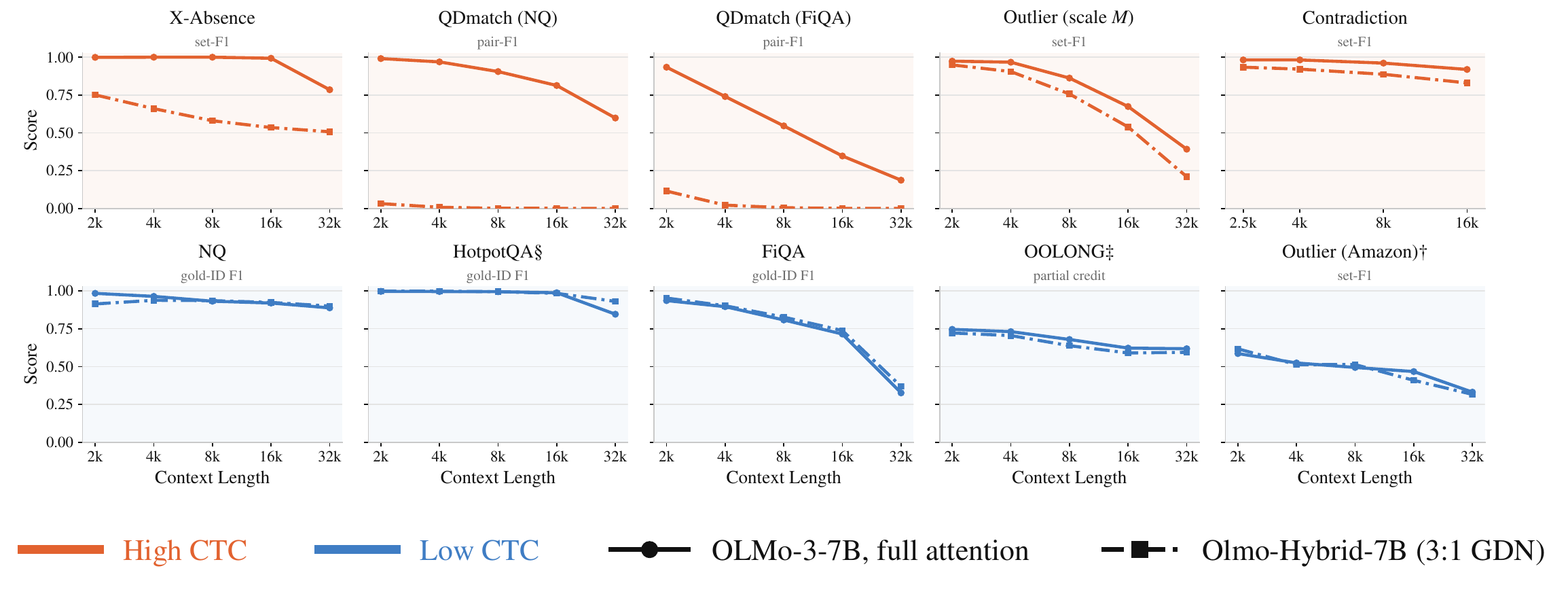}
\caption{\emph{Hybrid vs Full Models} We plot \bench-10 results for a full-attention (solid) version of OLMo-3-7B vs OLMo-3-Hybrid (dot-dashed). On low-\cname tasks both perform similarly, but on high-\cname tasks we observe consistent performance gaps.}

\label{fig:hybridcomp}
\end{figure}

\subsection{Full vs Hybrid}

\label{sec:fullvhybrid}

Beyond block-sparse attention, alternative architectures such as state-space-models (SSMs) \citep{Katharopoulos2020TransformersAR, Gu2023MambaLS} and adjacent approaches like \emph{gated delta net (GDN)} \citep{Yang2024GatedDN} have also enabled attention to reduce computational complexity. Specifically, these methods replace all-to-all token interactions with a recurrent state update, giving linear computation and constant memory. While this has shown promise, pure-GDN has, even at low-\cname, proven insufficient---language models like Qwen3.5 \citep{qwen3.5} have mixed GDN with full attention rather than replacing it.  


Strong performance of Qwen3.5 on long-context benchmarks \cite{Bai2024LongBenchVT}) and prior comparisons of full vs hybrid architectures \citep{Merrill2026OlmoHF} have suggested hybrid architectures to be comparable or even better than full attention, but does this free lunch apply to high-\cname tasks? To do a controlled test of this question, we compare OLMo-3-7B \citep{Ettinger2025Olmo3} vs OLMo-3-7B-Hybrid \citep{Merrill2026OlmoHF}, the closest open-weight comparison between hybrid and full architectures. Note that since Olmo3 by default uses sliding window attention, we adapt it to a full-attention version with a 100M-token continued-pretraining phase on Dolma3-Longmino data (0.1\% of its mid-training budget). OLMo-3-7B-Hybrid is already mid-trained natively on this data. 


\textbf{Hybrid Underperforms Full Attention on High-CTC Tasks.} We plot individual task results on the 10 \bench-10 tasks in Figure~\ref{fig:hybridcomp}. Similar to block-sparse attention, hybrid attention often leads to no degradation compared to full attention on low-\cname tasks (in fact doing better at 32k), but leads to large degradation on high-\cname. Note that these numbers are affected by OLMo models being generally weaker than Qwen3.5, and while overall trends are consistent, the gap varies by task, e.g. QDmatch (NQ) vs Outlier (wiki, Scale $M$). 

\subsection{Length Generalization} 
\label{sec:lengthgen}


Our work shows that even up to the relatively small scale of 32k tokens, high-\cname tasks already degrade greatly---scaling data sufficiently, especially for larger corpora, becomes increasingly important. But LCLM training is costly. Given these prohibitive costs, much work has sought to use short-context data to generalize to longer contexts (training only on short versions of tasks or short instruction-tuning data). Such work has successfully shown generalization on a variety of tasks past lengths used in fine-tuning \citep{Gao2024HowTT, Peng2023YaRNEC, Mehta2026RandomizedYI}. However, these results have predominantly been demonstrated only on low-\cname settings.



\textbf{High-\cname Tasks May Generalize Worse to Longer Contexts.} We evaluate 2k-32k-trained models at 64k and 128k on \bench-10. Note this does not require context extension as Qwen3.5-4B's native context is \~256K tokens. We plot results in Figure~\ref{fig:combined} (right) with full attention, showing percentage of 32k performance retained on the y-axis to increase comparability. At 128k context low-\cname{} ranges from 28\%-100\% while high-CTC ranges from 4\%-14\%---all high-\cname points fall below low-\cname points. We find these trends at only 128k tokens, and it's likely that this issue may become even more pronounced at larger corpus scales (e.g. 1 million+ tokens). Taken together with our prior results, this poses a conundrum: full attention is too costly to scale to large corpora, but many modeling decisions that make LCLMs tractable for larger corpora cause degradation on high-\cname tasks. 


\section{Analyses: Model Scale, Operation Difficulty}

We so far show that high-\cname often translates to tasks becoming much harder on larger corpora and how various ``free lunches'' that have been demonstrated on common tasks don't apply on high-\cname tasks. In this final section, we analyze how factors beyond complexity, both modeling (model family, model scale, \S\ref{sec:sparsityscaleresults}) and task-specific (oracle difficulty, \S\ref{sec:atomicoperations}) interplay with our previous findings.

\subsection{Role of Model Family and Scale} 

\label{sec:sparsityscaleresults}

Are our findings robust to different model scales and families? To answer this question, we sample two tasks: HotpotQA retrieval, one of the most commonly evaluated low-CTC tasks in \bench, and Contradiction, a representative high-\cname task. We train with three model families (OLMo-3-7B, Qwen3.5-4B, and Llama3.2-3B) \citep{dubey2024llama, Ettinger2025Olmo3}, as well as different model scales within Qwen3.5. We compare full vs block-sparse attention performance to see whether the trends align with our earlier finding.


\textbf{Block-Sparse Result Holds Across Model Families.}  
We report the performance gap between full and block-sparse attention across three sizes of Qwen 3.5, Llama, and OLMo in Figure~\ref{fig:model_scaling} (Left); individual results are reported in Figure~\ref{fig:family-lines} of Appendix~\ref{app:additional-exp}.
Across all families and scales, block-sparse attention causes little degradation on HotpotQA but substantial degradation on Contradiction. 
Comparing degrees of degration across three model sizes of Qwen-3.5  (0.8B, 2B, 4B) additionally reveal the pattern that smaller models degrades faster on larger input corpora.
This confirms our earlier finding that replacing full with block-sparse attention has a larger impact on high-CTC than low-CTC tasks.

\begin{figure}[t]
\centering
\begin{minipage}[b]{0.56\textwidth}
\centering
\includegraphics[width=\linewidth]{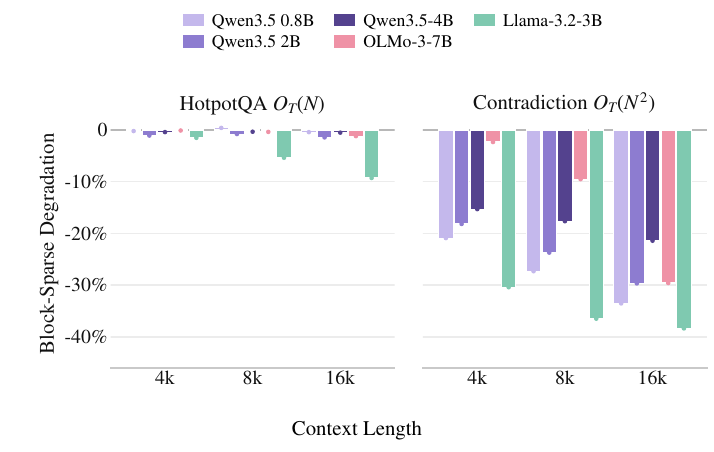}
\end{minipage}\hfill
\begin{minipage}[b]{0.42\textwidth}
\centering
\includegraphics[width=\linewidth]{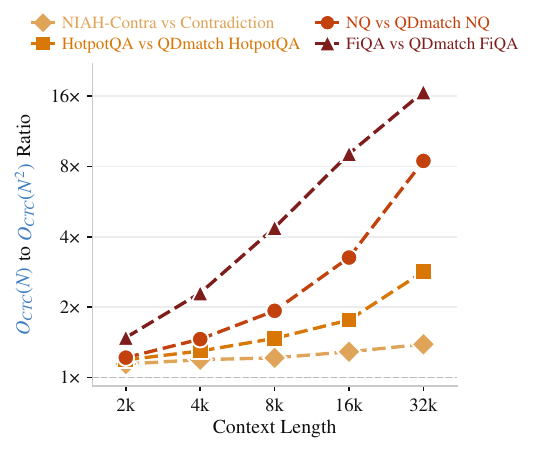}
\end{minipage}
\caption{ \textbf{(Left)} \emph{Model scale/family} Percentage of full-attention performance lost to block-sparse attention at 4k, 8k and 16k context, for three Qwen3.5 scales and two other model families, on a low-\cname{} (HotpotQA, \ctc{N}) and high-\cname{} (Contradiction, \ctc{N^2}) task. The gap is minimal on low-\cname{} and large and growing on high-\cname{}; within the Qwen3.5 family (shades of purple) smaller models degrade faster. \textbf{(Right)} \emph{Oracle Difficulty} For 4 task pairs with matched oracle operations (block-sparse attention), ratio of \ctc{N} to \ctc{N^2} task performance on the same corpus (pairs named in the legend). \textbf{Tasks with harder base operations (darker) see the ratio grow faster.}}
\label{fig:model_scaling}
\label{fig:atomic_difficulty}
\end{figure}

\subsection{Oracle Operation Difficulty}
\label{sec:atomicoperations}

CTC focuses on how the number of oracle calls required to solve a task scales with corpus size, rather than the difficulty of each oracle call itself. A natural question to ask is: How does oracle operation difficulty affect model performance on low- and high-CTC tasks?

To study this, we consider four \ctc{N} tasks, Contra-NIAH-Contra, HotpotQA, NQ, and FiQA, and rank them by relative difficulty, measured empirically by Qwen3.5-4B performance averaged across all evaluated context lengths. 
We pair each task with its corresponding \ctc{N^2} tasks---Contradiction, QDMatch HotpotQA, QDMatch NQ, QDMatch FiQA---and measure how much harder the \ctc{N^2} version is relative to its matched \ctc{N} task, i.e., the ratio of their performance at aech context length, under block-sparse attention.

\textbf{\cname Magnifies Difficulty Gaps Between Tasks.} Figure~\ref{fig:atomic_difficulty} (Right) shows that the performance gap between each \ctc{N^2} task and its matching \ctc{N}  tasks grows with context length. Importantly, the ordering of these gaps follows the ordering of oracle difficulty: tasks with harder oracle operations exhibit larger gaps as context length increases. This suggests that oracle difficulty and CTC interact systematically, and that relative degradation between tasks at shorter contexts may provide bounds on, or even predict degradation at longer contexts and higher CTC classes.

\section{Related Work}

\paragraph{Prior Corpus Reasoning Tasks} Corpus reasoning tasks (including examples with possibly high-\cname) have been studied for decades in computer science, from various low-\cname tasks in our suite (e.g. IR \cite{TREC2021}), to others like event ordering \citep{Chambers2014DenseEO}. Fields like Exploratory Data Analysis \citep{Tukey1962TheFO} (which studies visualization and  understanding of large data), though often focused on structured data, have explored engineering specialized pipelines  to understand unstructured data \citep{Shankar2024DocETLAQ}. Work in library sciences has also developed theories of interrelatedness between documents, most famously in the taxonomy of \citet{Tillett1991ATO}. More recently \citet{Zhang2025RecursiveLM}---which informally mentions a notion of task complexity related to CTC---includes an exhaustive pair generation task they find to be more challenging at scale for closed-source LLMs. Overall, while we assemble a large and new set of high-\cname tasks, we are not the first to propose tasks classifiable as high-\cname. That said, beyond existing notions \citep{Goldman2024IsIR}, we develop a new unified understanding of difficulty within CTC classes. To our knowledge this work is the first to identify LCLMs as uniquely suitable for high \cname tasks, and to connect task complexity to architectural decisions. Through \bench{}, our work also provides the most comprehensive testbed for how high-CTC task difficulty grows at scale.

\paragraph{Scalable Corpus Methods} Work applying language models to corpora has taken several directions. One path is to develop agentic scaffolds that combine efficient tools like dense retrievers \cite{Khattab2020ColBERTEA, Karpukhin2020DensePR} with LLMs. These are often hampered by fundamental challenges in retrieval \citep{Weller2025OnTT, Su2024BRIGHTAR, Wei2025BrowseCompAS}, but as shown by \cite{Zhang2025RecursiveLM} can be promising given the right scaffold and sufficient inference compute. Note that  the slowness of auto-regression (and the high per-query online cost, \citet{Zhang2025RecursiveLM}) make agentic systems intractable for high-complexity tasks---finding subtle contradictions via enumeration across $N$ documents may require generating on the order of $N^2$ tokens per query. Consequently, a complementary line of work---which we focus on in this work---seeks to develop architectures capable of ingesting full corpora end-to-end, often by making attention more efficient \citep{Acharya2024StarAE, Beltagy2020LongformerTL, Ivgi2022SLED, Lu2024TurboRAGAR, Xiao2025EfficientLCICL, Izacard2020LeveragingPR, Zaheer2020BigBT, Katharopoulos2020TransformersAR, Yang2024GatedDN}. Context extension \cite{Peng2023YaRNEC}, and length generalization \cite{Mehta2026RandomizedYI} also fall into these efforts. Our work helps develop a unified understanding of what these can / cannot achieve as a function of their computation. Importantly, our findings qualify this line of work, and reveal limitations obscured by the fact that these methods have primarily been developed on low-\cname tasks.

\section{Conclusion}

Our work focuses on clearly defining the challenges around high-\cname settings, and identifying the relationship between architecture design and \cname. While helping explain the success of cheaper linear cost methods, our investigation in particular encourages the community to avoid over-reliance on low-\cname evaluations and pay greater attention to high \cname tasks, which represent an interesting and impactful long-term open problem.

\section*{Acknowledgment}

We thank Ryan Wang, Karim Abdel Sadek, Jongho Park, Diane Tchuindjo, Omar Khattab, Suhas Kotha, Rhys Gould, Yaowen Ye, Yichuan Wang and other members of Jacob Steinhardt, Sewon Min, and Berkeley AI Research groups for discussion and feedback. This research was supported in part by ONR (N00014-26-1-2233), the NVIDIA Academic Grant Program, and gifts from Ai2 and Apple. This material is based upon work supported by the National Science Foundation Graduate Research Fellowship Program under Grant Numbers DGE2146752, DGE2637800, DGE2140739. Any opinions, findings, and conclusions or recommendations expressed in this material are those of the author(s) and do not necessarily reflect the views of the National Science Foundation. This project was also supported generously by compute from VESSL AI.


\bibliography{iclr2026_conference}

@article{Tillett1991ATO,
  title={A taxonomy of bibliographic relationships},
  author={Barbara B. Tillett},
  journal={Library Resources \& Technical Services},
  year={1991},
  volume={35},
  pages={150-158},
  url={https://api.semanticscholar.org/CorpusID:59748210}
}

@article{TREC2021,
  title={Overview of the TREC 2021 Deep Learning Track},
  author={Nick Craswell and Bhaskar Mitra and Emine Yilmaz and Daniel Fernando Campos and Jimmy J. Lin},
  journal={ArXiv},
  year={2025},
  volume={abs/2507.08191},
  url={https://api.semanticscholar.org/CorpusID:261242374}
}

@inproceedings{Hou2025CLERCAD,
  title={CLERC: A Dataset for U. S. Legal Case Retrieval and Retrieval-Augmented Analysis Generation},
  author={Abe Bohan Hou and Orion Weller and Guanghui Qin and Eugene Yang and Dawn J. Lawrie and Nils Holzenberger and Andrew Blair-Stanek and Benjamin Van Durme},
  booktitle={North American Chapter of the Association for Computational Linguistics},
  year={2025},
  url={https://api.semanticscholar.org/CorpusID:278664835}
}

@article{Su2024BRIGHTAR,
  title={BRIGHT: A Realistic and Challenging Benchmark for Reasoning-Intensive Retrieval},
  author={Hongjin Su and Howard Yen and Mengzhou Xia and Weijia Shi and Niklas Muennighoff and Han Wang and Hai-Bo Liu and Quan Shi and Zachary S. Siegel and Michael Tang and Ruoxi Sun and Jinsung Yoon and Sercan {\"O}. Arik and Danqi Chen and Tao Yu},
  journal={ArXiv},
  year={2024},
  volume={abs/2407.12883},
  url={https://api.semanticscholar.org/CorpusID:271270735}
}

@article{Robertson1976RelevanceWOBM25,
  title={Relevance weighting of search terms},
  author={Stephen E. Robertson and Karen Sp{\"a}rck Jones},
  journal={J. Am. Soc. Inf. Sci.},
  year={1976},
  volume={27},
  pages={129-146},
  url={https://api.semanticscholar.org/CorpusID:45186038}
}

@article{Khattab2020ColBERTEA,
  title={ColBERT: Efficient and Effective Passage Search via Contextualized Late Interaction over BERT},
  author={O. Khattab and Matei A. Zaharia},
  journal={Proceedings of the 43rd International ACM SIGIR Conference on Research and Development in Information Retrieval},
  year={2020},
  url={https://api.semanticscholar.org/CorpusID:216553223}
}

@article{Karpukhin2020DensePR,
  title={Dense Passage Retrieval for Open-Domain Question Answering},
  author={Vladimir Karpukhin and Barlas Oğuz and Sewon Min and Patrick Lewis and Ledell Yu Wu and Sergey Edunov and Danqi Chen and Wen-tau Yih},
  journal={ArXiv},
  year={2020},
  volume={abs/2004.04906},
  url={https://api.semanticscholar.org/CorpusID:215737187}
}

@article{Weller2025OnTT,
  title={On the Theoretical Limitations of Embedding-Based Retrieval},
  author={Orion Weller and Michael Boratko and Iftekhar Naim and Jinhyuk Lee},
  journal={ArXiv},
  year={2025},
  volume={abs/2508.21038},
  url={https://api.semanticscholar.org/CorpusID:280949957}
}

@article{Beltagy2020LongformerTL,
  title={Longformer: The Long-Document Transformer},
  author={Iz Beltagy and Matthew E. Peters and Arman Cohan},
  journal={ArXiv},
  year={2020},
  volume={abs/2004.05150},
  url={https://api.semanticscholar.org/CorpusID:215737171}
}

@article{Acharya2024StarAE,
  title={Star Attention: Efficient LLM Inference over Long Sequences},
  author={Shantanu Acharya and Fei Jia and Boris Ginsburg},
  journal={ArXiv},
  year={2024},
  volume={abs/2411.17116},
  url={https://api.semanticscholar.org/CorpusID:274280741}
}

@article{Ivgi2022SLED,
  title={Efficient Long-Text Understanding with Short-Text Models},
  author={Maor Ivgi and Uri Shaham and Jonathan Berant},
  journal={Transactions of the Association for Computational Linguistics},
  year={2022},
  volume={11},
  pages={284-299},
  url={https://api.semanticscholar.org/CorpusID:251224058}
}

@inproceedings{Lu2024TurboRAGAR,
  title={TurboRAG: Accelerating Retrieval-Augmented Generation with Precomputed KV Caches for Chunked Text},
  author={Songshuo Lu and Hua Wang and Yu Rong and Zhi Chen and Yaohua Tang},
  booktitle={Conference on Empirical Methods in Natural Language Processing},
  year={2024},
  url={https://api.semanticscholar.org/CorpusID:273233795}
}

@article{Xiao2025EfficientLCICL,
  title={Efficient Many-Shot In-Context Learning with Dynamic Block-Sparse Attention},
  author={Emily Xiao and Chin-Jou Li and Yilin Zhang and Graham Neubig and Amanda Bertsch},
  journal={ArXiv},
  year={2025},
  volume={abs/2503.08640},
  url={https://api.semanticscholar.org/CorpusID:276928367}
}

@article{Yen2024HELMETHT,
  title={HELMET: How to Evaluate Long-Context Language Models Effectively and Thoroughly},
  author={Howard Yen and Tianyu Gao and Minmin Hou and Ke Ding and Daniel Fleischer and Peter Izsak and Moshe Wasserblat and Danqi Chen},
  journal={ArXiv},
  year={2024},
  volume={abs/2410.02694},
  url={https://api.semanticscholar.org/CorpusID:273098808}
}

@article{Kocisk2017TheNR,
  title={The NarrativeQA Reading Comprehension Challenge},
  author={Tom{\'a}s Kocisk{\'y} and Jonathan Schwarz and Phil Blunsom and Chris Dyer and Karl Moritz Hermann and G{\'a}bor Melis and Edward Grefenstette},
  journal={Transactions of the Association for Computational Linguistics},
  year={2017},
  volume={6},
  pages={317-328},
  url={https://api.semanticscholar.org/CorpusID:2593903}
}

@article{izacardNQ,
  title={Few-shot Learning with Retrieval Augmented Language Models},
  author={Gautier Izacard and Patrick Lewis and Maria Lomeli and Lucas Hosseini and Fabio Petroni and Timo Schick and Jane A. Yu and Armand Joulin and Sebastian Riedel and Edouard Grave},
  journal={J. Mach. Learn. Res.},
  year={2022},
  volume={24},
  pages={251:1-251:43},
  url={https://api.semanticscholar.org/CorpusID:251371732}
}

@inproceedings{Yang2018HotpotQAAD,
  title={HotpotQA: A Dataset for Diverse, Explainable Multi-hop Question Answering},
  author={Zhilin Yang and Peng Qi and Saizheng Zhang and Yoshua Bengio and William W. Cohen and Ruslan Salakhutdinov and Christopher D. Manning},
  booktitle={Conference on Empirical Methods in Natural Language Processing},
  year={2018},
  url={https://api.semanticscholar.org/CorpusID:52822214}
}

@article{Lin2021PyseriniAP,
  title={Pyserini: A Python Toolkit for Reproducible Information Retrieval Research with Sparse and Dense Representations},
  author={Jimmy J. Lin and Xueguang Ma and Sheng-Chieh Lin and Jheng-Hong Yang and Ronak Pradeep and Rodrigo Nogueira and David R. Cheriton},
  journal={Proceedings of the 44th International ACM SIGIR Conference on Research and Development in Information Retrieval},
  year={2021},
  url={https://api.semanticscholar.org/CorpusID:235366815}
}

@article{Bertsch2025OolongEL,
  title={Oolong: Evaluating Long Context Reasoning and Aggregation Capabilities},
  author={Amanda Bertsch and Adithya Pratapa and Teruko Mitamura and Graham Neubig and Matthew R. Gormley},
  journal={ArXiv},
  year={2025},
  volume={abs/2511.02817},
  url={https://api.semanticscholar.org/CorpusID:282749185}
}

@article{Chen2026LongBenchPA,
  title={LongBench Pro: A More Realistic and Comprehensive Bilingual Long-Context Evaluation Benchmark},
  author={Ziyang Chen and Xing Wu and Junlong Jia and Chaochen Gao and Qingfang Fu and Debing Zhang and Songlin Hu},
  journal={ArXiv},
  year={2026},
  volume={abs/2601.02872},
  url={https://api.semanticscholar.org/CorpusID:284512608}
}

@article{Bai2024LongBenchVT,
  title={LongBench v2: Towards Deeper Understanding and Reasoning on Realistic Long-context Multitasks},
  author={Yushi Bai and Shangqing Tu and Jiajie Zhang and Hao Peng and Xiaozhi Wang and Xin Lv and Shulin Cao and Jiazheng Xu and Lei Hou and Yuxiao Dong and Jie Tang and Juanzi Li},
  journal={ArXiv},
  year={2024},
  volume={abs/2412.15204},
  url={https://api.semanticscholar.org/CorpusID:274859535}
}

@article{Zhang2025RecursiveLM,
  title={Recursive Language Models},
  author={Alex L. Zhang and Tim Kraska and Omar Khattab},
  journal={ArXiv},
  year={2025},
  volume={abs/2512.24601},
  url={https://api.semanticscholar.org/CorpusID:284350669}
}

@article{Hou2024BridgingLA,
  title={Bridging Language and Items for Retrieval and Recommendation},
  author={Yupeng Hou and Jiacheng Li and Zhankui He and An Yan and Xiusi Chen and Julian McAuley},
  journal={ArXiv},
  year={2024},
  volume={abs/2403.03952},
  url={https://api.semanticscholar.org/CorpusID:287635730}
}

@article{Izacard2020LeveragingPR,
  title={Leveraging Passage Retrieval with Generative Models for Open Domain Question Answering},
  author={Gautier Izacard and Edouard Grave},
  journal={ArXiv},
  year={2020},
  volume={abs/2007.01282},
  url={https://api.semanticscholar.org/CorpusID:220302360}
}

@article{Zaheer2020BigBT,
  title={Big Bird: Transformers for Longer Sequences},
  author={Manzil Zaheer and Guru Guruganesh and Kumar Avinava Dubey and Joshua Ainslie and Chris Alberti and Santiago Onta{\~n}{\'o}n and Philip Pham and Anirudh Ravula and Qifan Wang and Li Yang and Amr Ahmed},
  journal={ArXiv},
  year={2020},
  volume={abs/2007.14062},
  url={https://api.semanticscholar.org/CorpusID:220831004}
}

@inproceedings{Katharopoulos2020TransformersAR,
  title={Transformers are RNNs: Fast Autoregressive Transformers with Linear Attention},
  author={Angelos Katharopoulos and Apoorv Vyas and Nikolaos Pappas and Franccois Fleuret},
  booktitle={International Conference on Machine Learning},
  year={2020},
  url={https://api.semanticscholar.org/CorpusID:220250819}
}

@article{Gu2023MambaLS,
  title={Mamba: Linear-Time Sequence Modeling with Selective State Spaces},
  author={Albert Gu and Tri Dao},
  journal={ArXiv},
  year={2023},
  volume={abs/2312.00752},
  url={https://api.semanticscholar.org/CorpusID:265551773}
}

@article{Yang2024GatedDN,
  title={Gated Delta Networks: Improving Mamba2 with Delta Rule},
  author={Songlin Yang and Jan Kautz and Ali Hatamizadeh},
  journal={ArXiv},
  year={2024},
  volume={abs/2412.06464},
  url={https://api.semanticscholar.org/CorpusID:274598177}
}

@article{Wei2025BrowseCompAS,
  title={BrowseComp: A Simple Yet Challenging Benchmark for Browsing Agents},
  author={Jason Wei and Zhiqing Sun and Spencer Papay and Scott McKinney and Jeff Han and Isa Fulford and Hyung Won Chung and Alexandre Passos and William Fedus and Amelia Glaese},
  journal={ArXiv},
  year={2025},
  volume={abs/2504.12516},
  url={https://api.semanticscholar.org/CorpusID:277857238}
}

@article{Shankar2024DocETLAQ,
  title={DocETL: Agentic Query Rewriting and Evaluation for Complex Document Processing},
  author={Shreya Shankar and Aditya G. Parameswaran and Eugene Wu},
  journal={ArXiv},
  year={2024},
  volume={abs/2410.12189},
  url={https://api.semanticscholar.org/CorpusID:273374845}
}

@article{Tukey1962TheFO,
  title={The Future of Data Analysis},
  author={John W. Tukey},
  journal={Annals of Mathematical Statistics},
  year={1962},
  volume={33},
  pages={1-67},
  url={https://api.semanticscholar.org/CorpusID:122864799}
}

@misc{qwen3.5,
    title  = {{Qwen3.5}: Towards Native Multimodal Agents},
    author = {{Qwen Team}},
    month  = {February},
    year   = {2026},
    url    = {https://qwen.ai/blog?id=qwen3.5}
}

@article{Priem2022OpenAlexAF,
  title={OpenAlex: A fully-open index of scholarly works, authors, venues, institutions, and concepts},
  author={Jason Priem and Heather A. Piwowar and Richard Orr},
  journal={ArXiv},
  year={2022},
  volume={abs/2205.01833},
  url={https://api.semanticscholar.org/CorpusID:248512771}
}

@article{Jin2019PubMedQAAD,
  title={PubMedQA: A Dataset for Biomedical Research Question Answering},
  author={Qiao Jin and Bhuwan Dhingra and Zhengping Liu and William W. Cohen and Xinghua Lu},
  journal={ArXiv},
  year={2019},
  volume={abs/1909.06146},
  url={https://api.semanticscholar.org/CorpusID:202572622}
}

@inproceedings{kwon2023efficient,
  title={Efficient Memory Management for Large Language Model Serving with PagedAttention},
  author={Woosuk Kwon and Zhuohan Li and Siyuan Zhuang and Ying Sheng and Lianmin Zheng and Cody Hao Yu and Joseph E. Gonzalez and Hao Zhang and Ion Stoica},
  booktitle={Proceedings of the ACM SIGOPS 29th Symposium on Operating Systems Principles},
  year={2023}
}

@article{Goldman2024IsIR,
  title={Is It Really Long Context if All You Need Is Retrieval? Towards Genuinely Difficult Long Context NLP},
  author={Omer Goldman and Alon Jacovi and Aviv Slobodkin and Aviya Maimon and Ido Dagan and Reut Tsarfaty},
  journal={ArXiv},
  year={2024},
  volume={abs/2407.00402},
  url={https://api.semanticscholar.org/CorpusID:270870356}
}

@article{Fu2025AbsenceBenchLM,
  title={AbsenceBench: Language Models Can't Tell What's Missing},
  author={Harvey Yiyun Fu and Aryan Shrivastava and Jared Moore and Peter West and Chenhao Tan and Ari Holtzman},
  journal={ArXiv},
  year={2025},
  volume={abs/2506.11440},
  url={https://api.semanticscholar.org/CorpusID:279391603}
}

@article{Thakur2021BEIRAH,
  title={BEIR: A Heterogenous Benchmark for Zero-shot Evaluation of Information Retrieval Models},
  author={Nandan Thakur and Nils Reimers and Andreas Ruckl'e and Abhishek Srivastava and Iryna Gurevych},
  journal={ArXiv},
  year={2021},
  volume={abs/2104.08663},
  url={https://api.semanticscholar.org/CorpusID:233296016}
}

@article{Tchuindjo2026OBLIQBenchEO,
  title={OBLIQ-Bench: Exposing Overlooked Bottlenecks in Modern Retrievers with Latent and Implicit Queries},
  author={Diane Tchuindjo and Devavrat Shah and Omar Khattab},
  journal={ArXiv},
  year={2026},
  volume={abs/2605.06235},
  url={https://api.semanticscholar.org/CorpusID:288014448}
}

@article{Kwiatkowski2019NaturalQA,
  title={Natural Questions: A Benchmark for Question Answering Research},
  author={Tom Kwiatkowski and Jennimaria Palomaki and Olivia Redfield and Michael Collins and Ankur P. Parikh and Chris Alberti and Danielle Epstein and Illia Polosukhin and Jacob Devlin and Kenton Lee and Kristina Toutanova and Llion Jones and Matthew Kelcey and Ming-Wei Chang and Andrew M. Dai and Jakob Uszkoreit and Quoc V. Le and Slav Petrov},
  journal={Transactions of the Association for Computational Linguistics},
  year={2019},
  volume={7},
  pages={453-466},
  url={https://api.semanticscholar.org/CorpusID:86611921}
}

@article{Chambers2014DenseEO,
  title={Dense Event Ordering with a Multi-Pass Architecture},
  author={Nathanael Chambers and Taylor Cassidy and Bill McDowell and Steven Bethard},
  journal={Transactions of the Association for Computational Linguistics},
  year={2014},
  volume={2},
  pages={273-284},
  url={https://api.semanticscholar.org/CorpusID:1564278}
}

@article{Peng2023YaRNEC,
  title={YaRN: Efficient Context Window Extension of Large Language Models},
  author={Bowen Peng and Jeffrey Quesnelle and Honglu Fan and Enrico Shippole},
  journal={ArXiv},
  year={2023},
  volume={abs/2309.00071},
  url={https://api.semanticscholar.org/CorpusID:261493986}
}

@article{Mehta2026RandomizedYI,
  title={Randomized YaRN Improves Length Generalization for Long-Context Reasoning},
  author={Manas Mehta and Fangcong Yin and Greg Durrett},
  journal={ArXiv},
  year={2026},
  volume={abs/2606.23687},
  url={https://api.semanticscholar.org/CorpusID:289629978}
}

@article{Ettinger2025Olmo3,
  title={Olmo 3},
  author={Allyson Ettinger and Amanda Bertsch and Bailey Kuehl and David W. Graham and David Heineman and Dirk Groeneveld and Faeze Brahman and Finbarr Timbers and Hamish Ivison and Jacob Daniel Morrison and Jake Poznanski and Kyle Lo and Luca Soldaini and Matt Jordan and Mayee F. Chen and Michael Noukhovitch and Nathan Lambert and Pete Walsh and Pradeep Dasigi and Robert Berry and Saumya Malik and Saurabh Shah and Scott Geng and Shane Arora and Shashank Gupta and Taira Anderson and Teng Xiao and Tyler C. Murray and Tyler Romero and Victoria Graf and Akari Asai and Akshita Bhagia and Alexander Wettig and Alisa Liu and Aman Rangapur and Chloe Anastasiades and Costa Huang and Dustin Schwenk and Harsh M. Trivedi and Ian Magnusson and Jaron Lochner and Jiacheng Liu and Lester James Validad Miranda and Maarten Sap and Malia Morgan and Michaela Schmitz and Michal Guerquin and Michael Wilson and Regan Huff and Ronan Le Bras and Rui Xin and Rulin Shao and Sam Skjonsberg and Shannon Zejiang Shen and Shuyue Stella Li and Tucker Wilde and Valentina Pyatkin and William Merrill and Yapei Chang and Yuling Gu and Zhi-yuan Zeng and Ashish Sabharwal and Luke Zettlemoyer and Pang Wei Koh and Ali Farhadi and Noah A. Smith and Hannaneh Hajishirzi},
  year={2025},
  url={https://api.semanticscholar.org/CorpusID:283908770}, 
  journal={ArXiv},
}

@misc{dubey2024llama,
      title={The Llama 3 Herd of Models}, 
      author={Llama Team},
      year={2024},
      eprint={2407.21783},
      archivePrefix={arXiv},
      primaryClass={cs.AI}
}

@inproceedings{Gao2024HowTT,
  title={How to Train Long-Context Language Models (Effectively)},
  author={Tianyu Gao and Alexander Wettig and Howard Yen and Danqi Chen},
  booktitle={Annual Meeting of the Association for Computational Linguistics},
  year={2024},
  url={https://api.semanticscholar.org/CorpusID:273098476}
}

@article{Merrill2026OlmoHF,
  title={Olmo Hybrid: From Theory to Practice and Back},
  author={William Merrill and Yanhong Li and Tyler Romero and Anej Svete and Caia Costello and Pradeep Dasigi and Dirk Groeneveld and David Heineman and Bailey Kuehl and Nathan Lambert and Jacob Morrison and Luca Soldaini and Finbarr Timbers and Pete Walsh and Noah A. Smith and Hanna Hajishirzi and Ashish Sabharwal},
  journal={ArXiv},
  year={2026},
  volume={abs/2604.03444},
  url={https://api.semanticscholar.org/CorpusID:287199770}
}

@article{Gollapudi2026CanLM,
  title={Can Language Models Actually Retrieve In-Context? Drowning in Documents at Million Token Scale},
  author={Siddharth Gollapudi and Nilesh Gupta and Prasann Singhal and Sewon Min},
  year={2026},
  url={https://api.semanticscholar.org/CorpusID:289749122},
  journal={ArXiv},
}
\bibliographystyle{plainnat}

\appendix

\section{Limitations}
\claude{\label{app:limitations}}

While this work shows initial evidence for challenging properties of \cname tasks, there is still room for future work to expand on our analysis. Our paper, while supporting our hypotheses up to 32k tokens in-domain (128k tokens in our length generalization experiments) due to compute limitations does not examine larger corpus-scale settings beyond 128k tokens---note that for many high-CTC tasks performance already degrades greatly by 32k tokens and shows clear trends of degrading further. While we aim to get a comprehensive set (including 12 tasks representative of previously explored settings), there are still likely several low-\cname and high-\cname tasks our suite misses, and that we encourage future work to examine and reproduce our findings on. While this work helps define and understand high-\cname tasks, it only take initial steps towards solving them, and we are excited to see future work make further progress on high-CTC tasks.

\section{Discussion of Real High CTC Tasks}
\label{app:realism}

A key message of this work is to encourage more work on \ctc{N^2} and \ctc{NM} tasks, which represent very different architectural challenges at scale. While our investigation focuses largely on the technical and architectural aspects of this problem, we emphasize that while these tasks are under-explored, they \emph{are} realistic, and systems that can do such tasks over large corpora may open up various new valuable applications not previously possible, inspiring our current task suite. We below detail just a few applications and potential ideas for future work to investigate that could benefit from such systems at scale, connected to tasks from our suite:

\textbf{Contradiction / Redundancy Detection} (high-CTC tasks a) Motivating example: scientific literature. Given a large and growing body of scientific literature, a key and evolving question in determining new research and experiments is always \emph{what information does the community have?} Often, one will examine a specific research area at a local level, and given specific ideas may, with keyword search and other tools, try to determine what's been explored and not, or what's been verified or not, which is \ctc{N}. To determine the state of an entire field at a global level (e.g. all claims with the most contradictory evidence, claims with the most redundant evidence), allowing for systematic synthesis and resolution of such evidence, is \ctc{N^2}, and not currently tractable. Better corpus reasoning systems could allow such systematic analysis end-to-end for scientific literature, fact-checking, minimizing / tracking redundancy in long-text like books, and many other applications.

\textbf{Discovering long-tail categories / phenomena} (high-CTC tasks f, g) Motivating example: Monitoring LLM Generations. Language models can generate text at rapid scale, and for an LLM provider or other larger-scale user, every day a new corpus may get generated. For such instances, being able to faithfully look for the long tail of the least common events with harmful or other properties may be important to the safe deployment of such systems (e.g. \emph{What are the weirdest questions today?}). Doing so however is \ctc{NM} where $M$ is the number of possible phenomena (which in the long-tail case will be quite large), and so doing such a task at high-fidelity may currently be very challenging. Similar analyses could be applied for finding under-explored research areas / methodologies. 

\textbf{Structured Re-Organization of Large Corpora} (high-CTC task i) Corpora are often unstructured and, aside from some metadata, relatively unorganized. Being able to do analyses like sorting passages in a book by importance, or hierarchically re-organizing and grouping a corpus into a fashion reflecting the overall semantic structure (e.g. to reflect new trends in the corpus) can all be \ctc{NM} or \ctc{N^2}, and potentially provide valuable analyses or new ways of interacting with corpora.

\textbf{Cross-Corpus Comparison} (high-CTC tasks b, c, d, e) Often corpora don't exist as monoliths in isolation. We may be interested in understanding how corpora evolve over time  from largely similar distributions with high overlap (e.g. historical records, internet text, internal LLM agent traces from different live model checkpoints), or we may be interested in finding connections between unrelated corpora (for example two unrelated fields of science). These tasks are versatile, and in large part often end up being \ctc{N^2} as a function of the corpora being compared.

We believe that if systems which can handle such tasks become more scalable, given that the above are often difficult (or intractable) and may rely on a lot of manual work and heuristics with current systems, a large variety of applications may emerge from new methods here. Work towards building realistic benchmarks for these, as well as new architectures that can handle them, are valuable future directions. 

\section{Task Suite Data Details}
\label{app:data-details}

\providecommand{\ctc}[1]{\ensuremath{\mathcal{O}_T(#1)}}
\providecommand{\ex}[2]{#1~$\rightarrow$~\texttt{#2}}
\providecommand{\exblk}[1]{\multicolumn{2}{@{}l}{#1} \\}

\begin{table}[t]
\centering
\scriptsize
\setlength{\tabcolsep}{4pt}
\renewcommand{\arraystretch}{1.12}

\begin{tabular}{@{}
  >{\raggedright\arraybackslash}p{0.20\linewidth}
  >{\raggedright\arraybackslash}p{0.78\linewidth}
@{}}
\toprule
\textbf{Dataset} & \textbf{Example} \\
\midrule
\exblk{\textbf{\(\mathbf{O}(N)\): One-pass tasks}}
\midrule

NQ
  & \ex{\emph{``who sold out jesus for 30 pieces of silver''}}{[1]} \\
HotpotQA (bridge)
  & \ex{\emph{``What company did Rex Maughan aquire?''}}{[8, 9]} \\
NIAH-contra
  & \ex{\emph{``$\dots$a hypertonic solution of 14.4\% has an \textbf{ir}reversible ciliostatic
    effect''}}{[20]}, the one claim saying \emph{reversible} \\
BEIR SciFact
  & \ex{\emph{``0-dimensional biomaterials show inductive properties.''}}{[1]} \\
BEIR FiQA
  & \ex{\emph{``Where should I park my rainy-day / emergency fund?''}}{[2, 8, 10, 12, 19]} \\
MS MARCO
  & \ex{\emph{``what is the difference between fables and folktales? for kids''}}{[2]} \\
MS MARCO rerank
  & the same 20-passage pool, every passage scored $\rightarrow$
    \texttt{Ranking: [2], [1], [3], \dots} \\
OBLIQ
  & \ex{\emph{``$\dots$written in a similar style and voice to: The slowness and calmness of the
    lodge came with genuine inquisitiveness$\dots$''}}{[14, 23, 29]} \\
OOLONG
  & 10 labelled sentence pairs; \ex{\emph{``which of the labels is the most
    common?''}}{Label: incorrect} \\
Outlier (Amazon)
  & 20 reviews, 16 rated 4-star and 4 rated 5-star $\rightarrow$ \texttt{Outliers: 4, 7, 11, 17} \\
Outlier (wiki, fixed \(M\))
  & 57 chunks from 3 articles (43 / 11 / \textbf{3}) $\rightarrow$ \texttt{Outliers: 5, 17, 54} \\
Absence
  & 90 numbered sentences, 3 deleted from the copy $\rightarrow$ \texttt{["The interior is what",
    "In opposite chapels are", \dots]} \\

\midrule
\exblk{\textbf{Higher-complexity tasks}}
\midrule

Outlier (wiki, scale-\(k\))
  & 55 chunks from 10 articles (9 / 7 / 7 / $\dots$ / \textbf{3}) $\rightarrow$
    \texttt{Outliers: 26, 41, 42} \\
Grouping (OpenAlex)
  & \ex{\emph{``Cluster these 20 papers into 15 groups.''}}{\{"groups": [\{"doc\_ids":
    [4, 5, 12]\}, \dots]\}} \\
Contradiction
  & \texttt{[2]} \emph{``$\dots$has a reversible ciliostatic effect''} vs.\ \texttt{[12]}
    \emph{``$\dots$has an irreversible ciliostatic effect''} $\rightarrow$
    \texttt{[[2,12], [8,11], [13,17]]} \\
xabsence
  & 19 claims, 8 matched pairs, 3 with no twin in the other corpus $\rightarrow$
    \texttt{Unmatched: 1, 3, 14} \\
strmatch
  & \texttt{[1]} and \texttt{[11]} share \emph{``kossuth westward taco''} $\rightarrow$
    \texttt{[[1,11], [2,8], [15,19]]} \\
qdmatch (NQ)
  & \texttt{[7]} \emph{``when did 10 shilling note go out of circulation''} $\leftrightarrow$
    \texttt{[29]} $\rightarrow$ \texttt{[[7,29], [14,28], [19,40]]} \\
qdmatch (HPQA)
  & \texttt{[8]} \emph{``\,`Lost!' is a song by a British rock band formed in what year?''}
    $\leftrightarrow$ \texttt{[25]} and \texttt{[34]} \\
qdmatch (FiQA)
  & \texttt{[25]} \emph{``Pay off credit card debt or earn employer 401(k) match?''}
    $\leftrightarrow$ \texttt{[55]} \\
Reorder
  & 50 shuffled segments of \emph{Flatland} $\rightarrow$ \texttt{[31, 19, 12, 22, 37, \dots]} \\
Textgroups
  & noun counts $34{+}31{+}5 = 70$ and $13{+}17{+}40 = 70$ $\rightarrow$
    \texttt{[[1,11,12], [2,14,17]]} \\

\bottomrule
\end{tabular}

\caption{\textbf{One real example per task.} All content is verbatim from the evaluation sets,
elided with \texttt{\dots} where a passage is too long to show; document IDs are 1-indexed exactly
as the model sees them. Appendix~\ref{app:examples} shows each example in full, with the task
instruction, the surrounding corpus and the distractor types.
}
\label{tab:task-examples}
\end{table}

\subsection{Data Construction}
\label{app:construction}

\paragraph{Evaluation Set Construction} We evaluate at lengths up to 128k tokens (our code supports up to 10M tokens for several tasks). For each task we typically fix a canonical query set (usually 500 examples per set), and then will across different length evaluation sets within the task expand the distractor set. The documents and queries between train and eval sets are by default kept completely disjoint to avoid any sort of contamination.

\paragraph{Training Data Construction} Unless stated otherwise, each task is trained for one epoch on 20{,}000 examples split evenly across five context budgets (so 4{,}000 examples for 2k, 4k, 8k, 16k and 32k respectively, so the training distribution matches the evaluation uniformly. Training examples are drawn from the same generator as the evaluation examples for that task, and for the tasks with a published split (e.g. NQ, etc.), we use existing train/test divisions. OBLIQ is the sole exception: it has no usable training split and its training data is synthesized, though for evaluation we use its existing evaluation set with BM25 mined hard negatives.

\subsection{Tasks: Overview}

\subsubsection{\texorpdfstring{\ctc{N}}{OT(N)} tasks: one pass over the corpus}

\textbf{Factoid QA (NQ).} \emph{Gold-ID F1.} We use NaturalQuestions-Open (sourcing train and eval from existing train / eval set) \citep{izacardNQ}, with all documents --- gold, hard negative and filler --- drawn from the 100-word Wikipedia DPR corpus \citep{Karpukhin2020DensePR} served by pyserini \citep{Lin2021PyseriniAP}, so every document shares one surface format. For each question we BM25-search \citep{Robertson1976RelevanceWOBM25} for a passage containing an answer string (that becomes the gold), take top BM25 hits that do \emph{not} contain the answer as hard negatives, and fill the remaining slots with random wiki-DPR passages. By default we hold hard negatives to $\sim$10\% of the pool.

\textbf{Multi-hop QA (HotpotQA).} \emph{Gold-ID F1.} The bridge subset of HotpotQA \citep{Yang2018HotpotQAAD}, where answering requires composing two gold passages; the corpus is again wikipedia-100w with BM25 hard negatives. Because the answer needs a fixed (two) number of retrieval steps, this remains \ctc{N}. We source train examples from the standard train set and test examples from the validation set. We use \texttt{cross-encoder/ms-marco-MiniLM-L-6-v2} with a CE filter of 3 as well to remove false positives. And similar to NQ we hold hard negatives to $\sim$10\% of the pool.

\textbf{BEIR SciFact / FiQA.} \emph{Set-F1.} We use two BEIR \cite{Thakur2021BEIRAH} tasks: SciFact (scientific claim $\to$ abstract, single-gold) and FiQA (financial opinion QA, median 2 golds/query). Both mine distractors by BM25 from the dataset's own corpus. FiQA is sparsely judged, so plain BM25 distractors to filter unlabeled positives we score (query, candidate) pairs with a cross-encoder and keep a candidate only if it scores at least a margin below the gold, then fill the rest of the pool with random real corpus documents. FiQA's train-set is too small (5,500 examples) so we re-use queries across context lengths to get to 20,000. SciFact's existing train set was too small (809) to go to 20,000 total, so it has a smaller train set (4045 examples, each query used once per length). For SciFact the eval set is only 299 examples so our set doesn't go up to 500.  We use the same CE filter pipeline as HotpotQA to remove false positives.

\textbf{MS MARCO retrieval and reranking.} \emph{Set-F1; MRR@10 for docs with positive CE score} Passages come from the \texttt{msmarco-v1-passage} index. Hard negatives are taken from SentenceTransformers' precomputed BM25+dense mining and filtered by a margin on precomputed cross-encoder scores (same as above), so we don't re-run cross-encoder here; random fill is sampled as random pids from the 8.8M collection, which are byte-identical in format to gold and hard passages. The rerank variant reuses the identical pool but asks for the documents in ranked order based on cross-encoder score. 

\textbf{Needle contradiction (NIAH-contra).} \emph{Set-F1.} An \ctc{N} control derived from the \ctc{N^2} contradiction task: one member of a gold contradiction pair is promoted to the query (``\emph{find the document that contradicts this claim}'') and the other stays in the corpus as the needle. Every other claim, including the other gold pairs, is a distractor. Comparing this against contradiction isolates the effect of all-pairs search from the effect of PubMed claim text.

\textbf{OBLIQ retrieval.} \emph{Set-F1.} OBLIQ-Bench (\texttt{dianetc/OBLIQ-Bench}) poses subjective, long-form queries over its own corpora; we port it in-context by forcing all qrel positives into the example and filling with BM25-mined distractors from the same subset. Specifically, we use the twitter subset of the benchmark (``\emph{find tweets where users are implicitly insinuating that great powers quietly profit from sustained turmoil}\ldots{}''), satisfied by a \emph{set} of documents: median 5 golds per example, mean 9. The documents themselves are single tweets --- a median of 257 characters.

\emph{Synthetic training data.} As OBLIQ has no usable training split: the benchmark
has only a few hundred real queries per subset. We thus generate data as follows. For each seed document $D$ (taken from corpus):

\begin{enumerate}
\item An LLM (Qwen3-14B-Instruct) writes an oblique query $Q$ that $D$ satisfies, few-shot primed with real
(document, query) pairs (held out from final eval set).
\item An \emph{obliqueness} filter rejects any $Q$ with word-Jaccard overlap against $D$ above a
per-subset cap (0.55 for the writing subset, 0.35 elsewhere), so a keyword shortcut cannot separate
the golds --- matching real OBLIQ, whose queries are deliberately non-lexical.
\item BM25 retrieves candidate documents for $Q$ from the same subset.
\item A \emph{multi-gold judge} (also using Qwen3-8B) asks, for each top candidate, whether it \emph{also} satisfies $Q$. These either become additional golds or hard negatives.
\end{enumerate}

\textbf{Only training data is synthetic} --- the held-out evaluation set is the twitter subset of the benchmark (we use all subsets for generation of the training data). Our data is very initial and it's likely that much better data could be generated for this task.

\textbf{Oolong.} \emph{Partial-credit score.} A port of OOLONG \citep{Bertsch2025OolongEL}: a long context of labeled classification items, each tagged with a date and a user id, and a distributional question --- most/least common label, most frequent user, when label A overtook label B. We use the \emph{unlabeled} context variant, so the model must classify each item and then aggregate. Because the label set is fixed and each item is judged independently before a single aggregation step, this is \ctc{N}. 


\textbf{Absence.} \emph{Set-F1 over removed IDs.} Following AbsenceBench \citep{Fu2025AbsenceBenchLM}: the model sees a full numbered corpus, then a second copy with some elements deleted, and must name what is missing. We construct data with random Project Gutenberg passages, with sentence-level chunks: each document is deleted independently with probability $p$. Because the second copy preserves the original order, the two corpora are positionally aligned.

\textbf{Outlier detection (Amazon Reviews).} \emph{Set-F1.} Each corpus contains customer reviews from the Amazon Reviews 2023 corpus \citep{Hou2024BridgingLA}, shown as review title and body only. Examples are mixed 50/50 between two variants. In the \emph{rating} variant a majority of reviews share one star rating and 3 outliers carry a different one (\emph{``find the outliers with the least common rating''}); in the \emph{category} variant the majority share one product category and 3 outliers come from another (\emph{``find the outliers from the least common category''}), with categories taken from the corpus's own product-category labels.

\subsubsection{\texorpdfstring{\ctc{NM}}{OT(NM)} tasks: categorization}

\textbf{Outlier detection (Wikipedia).} \emph{Set-F1.} Each context is filled with 100-word chunks from a handful of Wikipedia articles with imbalanced counts; the answer is the ID list of chunks belonging to the article with the fewest chunks. We run two variants: \emph{scale-$M$}, where the number of source articles grows with $N$ ($M=N/5$ on average), and a \emph{fixed-$M$} control where the article count stays constant (at a value of 3) as $N$ grow.



\textbf{Grouping (OpenAlex).} \emph{Kendall Tau with true order labels} Given $N$ scientific abstracts, partition them into $M$ groups. Gold partitions come from the OpenAlex \citep{Priem2022OpenAlexAF} concept hierarchy: we sample only papers annotated down to level L3 (lower levels indicate more specific categories), then per example pick a level $L \in \{L0,\ldots,L3\}$ and sample $M$ distinct concept values at that level. The level sets the granularity---L0 gives coarse groups to L3 fine ones---and the model must infer from the corpus and $M$ what granularity is being asked for. Note that $M$ will naturally be larger for larger corpora with these heuristics.

\subsubsection{\texorpdfstring{\ctc{N^2}}{OT(N2)} tasks: all-pairs search}

\textbf{Cross-corpus absence (xabsence).} \emph{Set-F1.} Two corpora $A$ and $B$ share near identical document sets. Almost every document in one has a twin in the other; exactly $k$ are unmatched, and the model names these. The suite uses an \textbf{exact-copy} variant: a twin is an identical copy of its partner, and
documents are full PubMed abstracts rather than single sentences, but this task could be made more challenging with paraphrasing or other variants.

\textbf{Query-document matching (qdmatch).} \emph{Pair set-F1.} The retrieval analogue of contradiction for 3 retrieval tasks. Pools are drawn so that relevant questions, distractor questions (gold withheld) and distractor documents (questions withheld) are disjoint such that only three pairs of (question, document) are in the corpus. The model must find the sparse gold pairs among $N \times N$ combinations. We build three versions from three retrieval sources---NQ (1 gold/question), HotpotQA bridge (2 golds/question), and FiQA (median 2 golds/question)---so difficulty from the underlying retrieval problem varies while the search structure is fixed. Note that there are no hard negatives here, but the process to get train data and sources are the same as the original retrieval sets described above.

\textbf{Contradiction search (PubMed).} \emph{Pair set-F1} Each document is a single claim sentence from a PubMed abstract \citep{Jin2019PubMedQAAD}. Among $N$ documents, $K{=}3$ gold pairs $(d, d')$ are hidden: $d$ is a real PubMed sentence and $d'$ is a sentence, written by Qwen3-14B-Instruct, reporting what a \emph{different} study might have found such that the two cannot both be true (and prompted with the style of other claim pairs). A word-overlap threshold (Jaccard ${\leq}0.5$) drops any near-duplicates. Only one sentence per pair is LLM-written. Candidate sentences are filtered via regex to be self-contained and claim sentences, and fillers come from abstracts disjoint from every gold-source abstract in that example. While we manually inspect 100 examples for obvious artifacts, subtle LLM-generated annotation artifacts likely make this version of the task much easier than more realistic versions of contradiction search, which we encourage future work to investigate. 

\textbf{String matching (strmatch).} \emph{Pair set-F1.}  $N$ strings of $L$ words drawn from a Wikipedia vocabulary, and the task is to find every pair sharing a contiguous run of $\geq k$ words. Gold pairs share exactly one $k$-word run; hard negatives share a $(k{-}1)$-word run. This task has no semantic content at all, so the only thing that scales is the number of comparisons.

\textbf{Reordering.} \emph{Kendall-$\tau$.} $N$ consecutive sentence segments of a single Project Gutenberg book, presented in random order, with the task of recovering the original ordering as a list of document IDs. Because sorting requires comparisons between arbitrary pairs of segments, we label it \ctc{N^2} (unlike number sorting we can't assume a strict global positional prior). 

\subsubsection{Beyond \texorpdfstring{\ctc{N^2}}{OT(N2)}}

\textbf{Textgroups.} \emph{Group set-F1.} $N$ chunks (sourced from Project Gutenberg), each carrying a feature (how many nouns / verbs / adjectives it contains, or how often a chosen connective appears), and the task is to find every triple whose feature values sum to a target $T$. Because $G{=}3$, the construction can both plant exactly $K$ triples and verify by brute force that no others exist. This is the one task in the suite where the per-document oracle operation is itself nontrivial (counting) \emph{and} the CTC is greater than quadratic. We include it as an example of tasks that can go beyond \ctc{N^2}.

\section{Data Statistics / Evaluation}
\label{app:data-stats}

Table~\ref{tab:data-stats} lists, for every task in the suite, the corpus it is built from, its
\cname{} class, the number of documents present at each context rung, and the number
of examples it is evaluated on. Each example carries \texttt{documents} (a list of \{title, text\}),
\texttt{queries}, \texttt{answers}, and \texttt{gold\_doc\_indices}. Document counts are measured
from the evaluation files. Document length is given in characters.

\textbf{Evaluation} We typically serve everything with vLLM \cite{kwon2023efficient} (we use a modified version for block-sparse inference), where block-sparse attention is implemented by masking with respect to marker tokens. For generation we use greedy decoding.

\begin{table}[htbp]
\centering
\scriptsize
\setlength{\tabcolsep}{3.2pt}
\begin{tabular}{l l c l r r r r r r r}
\toprule
& & & & \multicolumn{5}{c}{\textbf{Documents per example ($N$)}} & & \\
\cmidrule(lr){5-9}
\textbf{Task} & \textbf{Source corpus} & \textbf{\cname} & \textbf{Metric}
 & 2k & 4k & 8k & 16k & 32k & \textbf{Chars/doc} & \textbf{Eval} \\
\midrule
NQ & Wikipedia 100w (DPR) & \ctc{N} & gold-ID F1 & 11 & 23 & 48 & 104 & 208 & 636 & 500 \\
HotpotQA (bridge) & Wikipedia 100w (DPR) & \ctc{N} & gold-ID F1 & 17 & 36 & 72 & 144 & 296 & 432 & 500 \\
NIAH-contra & PubMed claims & \ctc{N} & gold-ID F1 & 40 & 86 & 180 & 365 & 740 & 60 & 500 \\
BEIR SciFact & SciFact abstracts & \ctc{N} & gold-ID F1 & 5 & 10 & 21 & 43 & 88 & 1517 & 300 \\
BEIR FiQA & FiQA-2018 posts & \ctc{N} & gold-ID F1 & 8 & 19 & 40 & 82 & 166 & 835 & 500 \\
MS MARCO & msmarco-v1-passage & \ctc{N} & gold-ID F1 & 23 & 46 & 93 & 187 & 374 & 337 & 500 \\
MS MARCO rerank$\ddagger$ & msmarco-v1-passage & \ctc{N} & MRR@10 & 20 & 40 & 70 & 100 & 376 & 336 & 500 \\
OBLIQ & OBLIQ-Bench posts & \ctc{N} & gold-ID F1 & 3 & 6 & 30 & 19 & 30 & 4196 & 126 \\
OOLONG$\S$ & labeled item stream & \ctc{N} & partial credit & 41 & 83 & 153 & 337 & 616 & 154 & 500 \\
Outlier (Amazon) & Amazon Reviews 2023 & \ctc{N} & set-F1 & 20 & 40 & 80 & 160 & 320 & 494 & 500 \\
Outlier (wiki, fix-$M$) & Wikipedia 100w & \ctc{N} & set-F1 & 14 & 28 & 57 & 111 & 220 & 612 & 500 \\
Absence & Project Gutenberg & \ctc{N} & set-F1 & 90 & 180 & 360 & 720 & -- & 146 & 500 \\
\midrule
Outlier (wiki, scale-$k$) & Wikipedia 100w & \ctc{NM} & set-F1 & 14 & 28 & 57 & 111 & 220 & 612 & 500 \\
Grouping & OpenAlex abstracts & \ctc{NM} & pairwise-F1 & 10 & 21 & 43 & 88 & 176 & 893 & 500 \\
Textgroups & synthetic passages & \ctc{N^3} & group-F1 & 11 & 24 & 50 & 103 & 210 & 1182 & 500 \\
Contradiction & PubMed claims & \ctc{N^2} & set-F1 & 56 & 92 & 187 & 379 & 762 & 148 & 500 \\
Cross-corpus absence & PubMed claims & \ctc{N^2} & set-F1 & 39 & 81 & 165 & 333 & 669 & 139 & 500 \\
String matching & synthetic strings & \ctc{N^2} & set-F1 & 38 & 82 & 170 & 350 & 700 & 81 & 500 \\
QDmatch (NQ) & Wikipedia 100w (DPR) & \ctc{N^2} & pair-F1 & 18 & 40 & 84 & 174 & 356 & 338 & 500 \\
QDmatch (HotpotQA) & Wikipedia 100w (DPR) & \ctc{N^2} & pair-F1 & 18 & 40 & 84 & 174 & 356 & 272 & 500 \\
QDmatch (FiQA) & FiQA-2018 posts & \ctc{N^2} & pair-F1 & 10 & 26 & 58 & 122 & 248 & 548 & 500 \\
Reordering & Project Gutenberg & \ctc{N^2} & Kendall tau & 12 & 27 & 57 & 116 & -- & 581 & 500 \\
\bottomrule
\end{tabular}
\caption{\textbf{Corpus statistics for every task in the evaluation suite} Column headings 2k--32k are approximate token targets per task; the entries are the median number
of documents actually present in an example of that rung, measured from the evaluation files the
reported results were graded on. Chars/doc is the median document length in characters at the
deepest available rung. }
\label{tab:data-stats}
\end{table}

\section{Additional Experiments}\label{app:additional-exp}

\subsection{Model Scale}

\begin{figure}[t]
\centering
\includegraphics[width=\textwidth]{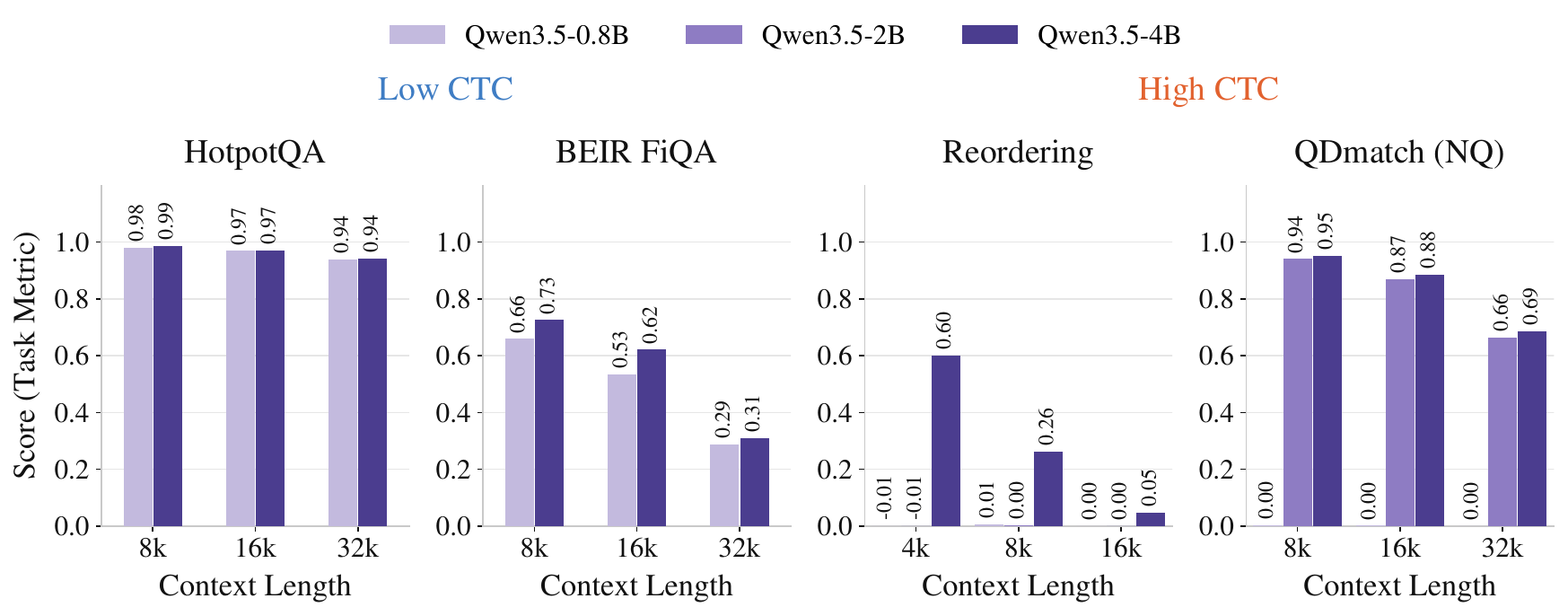}
\caption{Full model performance on several tasks and Qwen3.5 model scales. Some high-CTC tasks have minimum model scales necessary to learn effectively.}
\label{fig:scaling-lines}
\end{figure}

We include a more detailed plot with some different dense and block-sparse attention values for our model scale comparison (Figure~\ref{fig:scaling-lines}). This includes dense attention model scale results on Reordering and QDMatch (NQ), where we find a trend of smaller model scales sometimes being unable to learn challenging high CTC tasks. 

\subsection{Model Family} 

\begin{figure}[t]
\centering
\includegraphics[width=\textwidth]{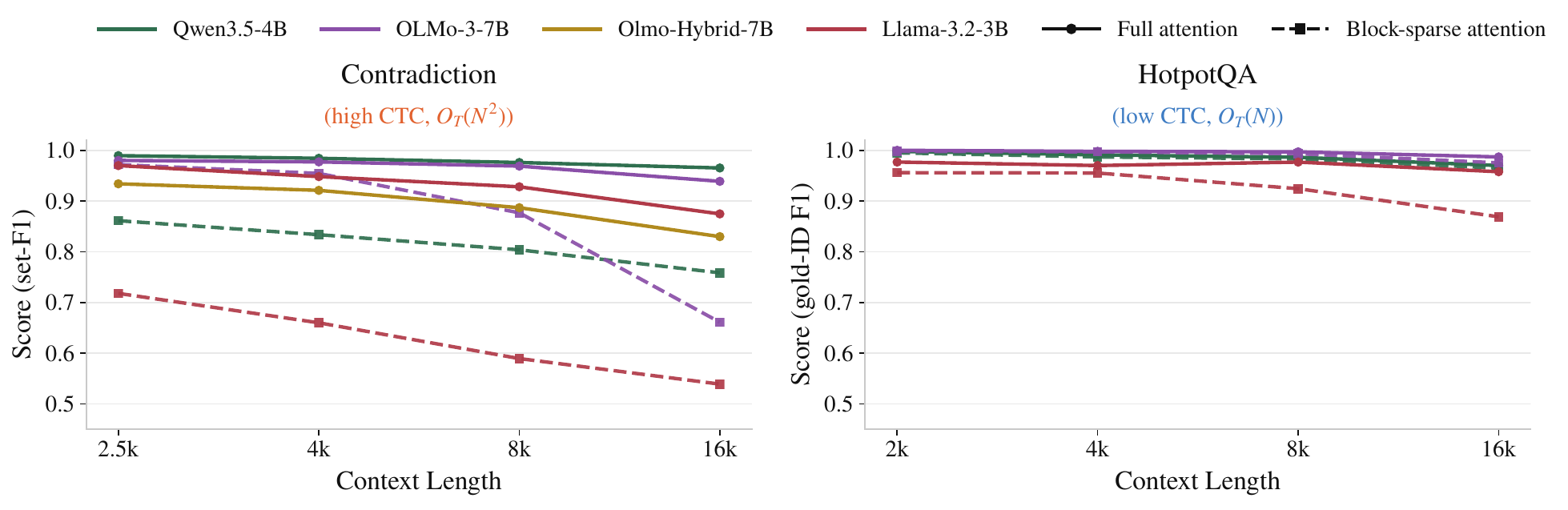}
\caption{Full vs block-sparse attention with a few different model families.}
\label{fig:family-lines}
\end{figure}

We in Figure~\ref{fig:family-lines} show a few exact model family performance numbers with both block-sparse and full attention. Note that block-sparse struggles greatly on contradiction across model families.

\section{Training Hyperparameters}
\label{app:hyperparams}

All CTC Suite results are full-parameter fine-tunes of a base checkpoint, trained with
\texttt{olmo-core}. Every arm of a comparison --- full attention, block-sparse, and
block-sparse with mask mixing --- is trained from the same base with the same recipe, so the
only difference between arms is the attention mask used during training. Defaults shared across
all settings:

\textbf{Optimizer:} AdamW with skip-step (a step whose loss or gradient norm is a large
outlier is skipped rather than applied), lr $5\!\times\!10^{-5}$, $\beta = (0.9, 0.95)$, weight
decay 0, gradient clipping at global norm 1.0.
\textbf{Schedule:} linear warmup over the first 3\% of steps, then linear decay to 0.
\textbf{Epochs / batch:} 1 epoch, global batch 8 sequences, 1 instance per micro-batch.
\textbf{Precision:} bf16 parameters with fp32 gradient reduction, FSDP for data parallelism,
activation checkpointing on.
\textbf{Packing:} examples are \emph{not} packed; one example per sequence, so per-example
sequence lengths are stable across runs and document boundaries stay aligned with chunk boundaries.

\paragraph{Mask mixing.}

We apply mask-mixing as a \emph{curriculum} from $0.8$ to $0$ rather than a fixed $p$, finding this to work well empirically. Evaluation always uses the block-sparse mask with no mixing, so a mask-mixing result reflects pure block-sparse inference and differs from the pure-block-sparse arm only in how it was trained.


\section{GPU Resources / Compute}
\label{app:gpus}

The experiments for our investigation involved roughly 3k A100 hours and 1k H200 hours (32 CPU cores, 300GB RAM). We expect that with 8xH100 GPUs, the results in the paper Figure~\claude{\ref{fig:multitask_fig4}} would take under 500 hours to reproduce.


\section{Task Examples}
\label{app:examples}


We show one real example for every task in the suite (Table~\claude{\ref{tab:eval-datasets}}), we omit full corpus text omitted for readability.

\subsection{\texorpdfstring{\ctc{N}}{OT(N)} tasks}

\subsubsection{NQ (factoid retrieval)}

\textbf{Instruction:} \emph{``Use the given documents to identify which document is most relevant to
answering the question. Write your answer in the following format: Relevant Document: [id]''}\\
\textbf{Query:} \emph{``who sold out jesus for 30 pieces of silver''}\\
\textbf{Context (3 of 20 documents shown):}
\begin{quote}\small
\texttt{[1]} \emph{(gold)} ``Thirty pieces of silver was the price for which Judas Iscariot betrayed
Jesus, according to an account in the Gospel of Matthew 26:15 $\dots$''\\
\texttt{[2]} \emph{(BM25 hard negative)} ``Butuan Silver Paleograph. The Butuan Silver Palaeograph,
also known as the `Butuan Silver Strip', is a piece of metal with inscriptions $\dots$''\\
\texttt{[3]} \emph{(BM25 hard negative)} ``Phantom Stranger $\dots$ of betraying him. As the Spectre
is about to attack the Stranger, a Mysterious Voice sends him off $\dots$ As payment for what
occurred $\dots$''\\
\texttt{[$\cdot$ 17 more documents omitted]}
\end{quote}
\textbf{Gold doc ID:} \texttt{1}\quad\textbf{Answer string (not the target):} \emph{``Judas Iscariot''}

\subsubsection{HotpotQA (bridge, multi-hop retrieval)}

\textbf{Instruction:} \emph{``$\dots$identify which documents are relevant $\dots$ Relevant
Documents: [id1], [id2]''}\\
\textbf{Query:} \emph{``What company did Rex Maughan aquire?''} \emph{(question text verbatim,
including the source typo)}\\
\textbf{Context (4 of 20 documents shown):}
\begin{quote}\small
\texttt{[8]} \emph{(gold-1)} ``Forever Living Products International, Inc.\ (FLPI) is an American
privately-held multi-level marketing (MLM) company based in Scottsdale, Arizona $\dots$''\\
\texttt{[9]} \emph{(gold-2)} ``Rex G. Maughan is an American businessman. He is the founder,
president, and chief executive officer of Forever Living Products $\dots$''\\
\texttt{[1]} \emph{(hard negative --- name collision)} ``Alex Maughan (born April 24, 1995) is an
American rugby union player who plays in the front row for the United States men's national team
$\dots$''\\
\texttt{[2]} \emph{(hard negative --- same article, wrong chunk)} ``Ruth graduated from BYU with a
degree in elementary education, she and Rex moved to Arizona $\dots$''\\
\texttt{[$\cdot$ 16 more documents omitted]}
\end{quote}
\textbf{Gold doc IDs:} \texttt{[8, 9]}\quad\textbf{Answer string:} \emph{``Aloe Vera of America''}

\subsubsection{NIAH-contradiction (\texorpdfstring{\ctc{N}}{OT(N)} control for contradiction)}

\textbf{Instruction:} \emph{single-document retrieval, as in NQ.}\\
\textbf{Query:} \emph{``Find the document that directly contradicts the following claim: The National
Lung Screening Trial, performed mainly in academic medical centers, showed that cancer mortality can
be reduced with computed tomography (CT) screening compared with chest radiography in high-risk
patients.''}\\
\textbf{Context (4 of 40 documents shown; the 2k rung):}
\begin{quote}\small
\texttt{[23]} \emph{(needle)} ``A large-scale randomized trial indicated that, for individuals at
elevated risk, chest X-rays were more effective than CT scans in lowering cancer-related deaths.''\\
\texttt{[1]} \emph{(filler)} ``Thirteen focus group discussions involving a total of 97 informants
were conducted.''\\
\texttt{[2]} \emph{(filler)} ``TRIM47 expression levels were found to be significantly increased in PC
compared to benign tissues by both immunohistochemistry and qRT-PCR $\dots$''\\
\texttt{[3]} \emph{(filler)} ``Free cortisol was significantly elevated at 8:00 to 8:30 hours in the
high job strain group but not at later times of the day or evening.''\\
\texttt{[$\cdot$ 36 more documents omitted, drawn from PubMed abstracts disjoint from the needle's
source abstract]}
\end{quote}
\textbf{Gold doc ID:} \texttt{23}.

\subsubsection{BEIR SciFact}

\textbf{Instruction:} \emph{single/multi-document retrieval.}\\
\textbf{Query (a scientific claim):} \emph{``0-dimensional biomaterials show inductive
properties.''}\\
\textbf{Context (3 of 22 abstracts shown):}
\begin{quote}\small
\texttt{[1]} \emph{(gold)} ``New opportunities: the use of nanotechnologies to manipulate and track
stem cells. Nanotechnologies are emerging platforms that could be useful in measuring,
understanding, and manipulating $\dots$''\\
\texttt{[2]} \emph{(BM25 negative)} ``The spectrum of retinopathy in adults with Plasmodium
falciparum malaria $\dots$''\\
\texttt{[3]} \emph{(BM25 negative)} ``Inheritance of coronary artery disease in men: an analysis of
the role of the Y chromosome $\dots$''\\
\texttt{[$\cdot$ 19 more documents omitted]}
\end{quote}
\textbf{Gold doc ID:} \texttt{1}

\subsubsection{BEIR FiQA (cross-encoder--cleaned negatives)}

\textbf{Instruction:} \emph{multi-document retrieval.}\\
\textbf{Query:} \emph{``Where should I park my rainy-day / emergency fund?''}\\
\textbf{Context (3 of 20 documents shown; 5 golds in this example):}
\begin{quote}\small
\texttt{[2]} \emph{(gold)} ``I would suggest your local credit union or local bank for security and
liquidity. Liquidity is probably the most important issue for a emergency fund.''\\
\texttt{[8]} \emph{(gold)} ``This is probably a good time to note that credit is not a liquid asset,
and not an emergency fund. Credit can be revoked or denied at any time $\dots$''\\
\texttt{[10]} \emph{(gold)} ``First off, you generally want to park your emergency fund somewhere
that is `safe', meaning something that is not subject to market fluctuations $\dots$''\\
\texttt{[$\cdot$ 17 more documents omitted: cross-encoder-surviving hard negatives ($\sim$10\% of
slots) plus random real corpus documents]}
\end{quote}
\textbf{Gold doc IDs:} \texttt{[2, 8, 10, 12, 19]}. FiQA is sparsely judged, so candidates scoring
close to the gold under the cross-encoder are dropped rather than used as negatives --- they are
likely unlabeled positives.

\subsubsection{MS MARCO (retrieval)}

\textbf{Instruction:} \emph{single-document retrieval.}\\
\textbf{Query:} \emph{``what is the difference between fables and folktales? for kids''}\\
\textbf{Context (3 of 20 passages shown):}
\begin{quote}\small
\texttt{[2]} \emph{(gold)} ``Folktale vs Fable. Folktales and fables can be understood as two
different types of stories that show a difference between them $\dots$''\\
\texttt{[1]} \emph{(hard negative)} ``Your job is to make sense of the fact or example in the context
of the overall main idea that is being conveyed $\dots$''\\
\texttt{[3]} \emph{(random fill)} ``Cascamite is a waterproof glue and is probably the must effective
glue of all $\dots$''\\
\texttt{[$\cdot$ 17 more documents omitted]}
\end{quote}
\textbf{Gold doc ID:} \texttt{2}. Hard and random passages are drawn from the same index, so they are
format-identical: there is no stylistic cue distinguishing a mined negative from a random one.

\subsubsection{MS MARCO reranking}

\textbf{Instruction:} \emph{``Rank the documents by how relevant each is to the question, from most to
least relevant. Output the document IDs in ranked order. Write your answer in the following format:
Ranking: [id1], [id2], [id3], \dots''}\\
\textbf{Query:} \emph{``what is the difference between fables and folktales? for kids''}\\
\textbf{Context (5 of 23 passages shown; the 2k rung. Cross-encoder scores not in the actual prompt):}
\begin{quote}\small
\texttt{[3]} \emph{(CE $+8.04$; also the MS MARCO qrel positive)} ``Folktale vs Fable. Folktales and
fables can be understood as two different types of stories that show a difference between them.
Mostly, folktales and fables are passed on from one generation to another orally $\dots$''\\
\texttt{[17]} \emph{(CE $+4.60$)} ``A fable differs from a parable in that the latter excludes animals,
plants, inanimate objects, and forces of nature as actors that assume speech $\dots$''\\
\texttt{[12]} \emph{(CE $+4.40$)} ``A fable is a moral tale that often features animal characters. We
often associate fables with the master of them all, Aesop $\dots$''\\
\texttt{[9]} \emph{(CE $-9.79$)} ``A mood disorder is a mental health class that health professionals
use to broadly describe all types of depression and bipolar disorders $\dots$''\\
\texttt{[19]} \emph{(CE $-11.30$, the lowest in the pool)} ``Geology 1003 with Weaver at University of
Oklahoma $\dots$''\\
\texttt{[$\cdot$ 18 more passages omitted]}
\end{quote}
\textbf{Target (top-10):} \texttt{Ranking: [3], [17], [1], [12], [9], [22], [13],
[16], [14], [5]}\\

\subsubsection{OBLIQ (subjective retrieval, twitter subset)}

\textbf{Instruction:} \emph{multi-document retrieval.}\\
\textbf{Query:} \emph{``Find tweets where users are implicitly insinuating that great powers quietly
profit from sustained turmoil in West Asia, suggesting that disruptions at key maritime chokepoints
and damage to energy facilities maintain elevated fuel costs and tilt market share toward them, while
questioning which actors truly profit from an extended standoff.''}\\
\textbf{Context (3 of 33 documents shown; 2k rung):}
\begin{quote}\small
\texttt{[19]} \emph{(gold)} ``Yes\ldots{} So what happens if all of that Middle Eastern infrastructure
gets destroyed over the next year or so including Iran's? Who benefits in terms of oil sales? Mostly
the US and Russia.''\\
\texttt{[27]} \emph{(gold)} ``Who stands to gain and who stands to lose if the Strait of Hormuz stays
closed?''\\
\texttt{[3]} \emph{(distractor)} ``Strong conviction on the spike! Brent's holding near \$103 spot
right now with Hormuz disruptions from the Iran strikes. Jumping to \$119 today or \$140 by Monday
would demand major sustained\ldots{}''\\
\texttt{[$\cdot$ 30 more documents omitted, BM25-mined from the twitter subset's 500-document pool]}
\end{quote}
\textbf{Gold doc IDs:} \texttt{[4, 19, 27, 30, 32]}

\subsubsection{OOLONG (aggregation over labeled items)}

\textbf{Instruction:} \emph{``Read the data below and answer the question. Compute the exact answer by
analyzing every item; do not guess or approximate.''}\\
\textbf{Context (rendered verbatim as one stream, not as numbered documents; 2k rung, 59 items, one
per line):}
\begin{quote}\small
``The following lines contain 59 general-knowledge questions, one per line. Each question has an
answer that can be described as one of 6 categories: `abbreviation', `entity', `human being',
`numeric value', `location', `description and abstract concept'.\\
You will be asked to answer questions about the aggregate label statistics across all 59 examples in
this dataset. Do not try to guess, estimate, or approximate the result. Calculate the exact answer
given these datapoints.\\
\texttt{Date: Oct 24, 2022 || User: 34487 || Instance:} \emph{Where is Inoco based ?}\\
\texttt{Date: Jul 14, 2022 || User: 11784 || Instance:} \emph{Where did the term `` 86 '' come from ?}\\
\texttt{Date: Feb 23, 2023 || User: 57329 || Instance:} \emph{What international sport was originally
called hurley ?}\\
\texttt{Date: Feb 09, 2024 || User: 24838 || Instance:} \emph{What is SVHS ?}\\
\texttt{Date: Apr 17, 2025 || User: 60511 || Instance:} \emph{What is HDLC ?}\\
\emph{[$\cdot$ 54 more items omitted]}''
\end{quote}
\textbf{Query:} \emph{``For the following question, only consider the subset of instances that occur in
April of any year. Among instances occuring in April, which of the labels is the most common? Give your
final answer in the form `Label: answer' $\dots$''}\\
\textbf{Answer:} \texttt{abbreviation}

\subsubsection{Outlier (Amazon Reviews)}

\textbf{Instruction:} \emph{``You are given a list of product reviews. Most share a common attribute
(star rating or product category); a few are outliers with a different value. First state what the
majority attribute is and what the outlier attribute is, then list the 1-indexed document IDs of the
outliers $\dots$ Outliers: [id1], [id2], \dots''}\\
\textbf{Query:} \emph{``Can you find outliers with the least common rating in this data?''}\\
\textbf{Context (3 of 20 reviews shown; ratings and categories are withheld from the model ---
only title and body are rendered):}
\begin{quote}\small
\texttt{[4]} \emph{(\textbf{outlier}, 5-star)} \emph{``Works great, no errors on memtest''} --- ``This
RAM comes in 2 sticks of 4gb $\dots$ and works perfectly in my old PC. Ran memtest with 0 errors
$\dots$''\\
\texttt{[11]} \emph{(\textbf{outlier}, 5-star)} \emph{``Great Case at a Fantastic Price''} --- ``I
never realized Amazon had their own line of products, AmazonBasics! $\dots$ The case is the perfect
size for my new Canon Powershot Camera $\dots$''\\
\texttt{[1]} \emph{(majority, 4-star)} \emph{``Light''} --- ``Works well. Can be kind of heavy
depending on the size of your camera. Easy to install and use $\dots$''\\
\texttt{[17 more reviews omitted]}
\end{quote}
\textbf{Target:} \emph{``Most reviews are 4-star ratings and the outliers are 5-star reviews.''}
followed by \texttt{Outliers: [4], [7], [11], [17]}. Ground truth comes from structured metadata only
(the star-rating field), never from an LLM label. The attribute set is small and fixed --- five star
ratings, or a closed category list --- which is what keeps this variant \ctc{N}: there is no category
to discover.

\subsubsection{Outlier (Wikipedia, fixed M)}

The \ctc{N} control for the scale-$k$ outlier task: identical format and generator, but the number of
source articles $M$ is pinned as $N$ grows instead of scaling with it.\\
\textbf{Query:} \emph{``Can you find passages that are about a different topic than the rest of these
passages?''}\\
\textbf{Context (3 of 14 chunks shown at the 2k rung; article titles are withheld):}
\begin{quote}\small
\texttt{[2]} \emph{(\textbf{outlier})} ``1987, and 1989. Several notable players played for Sheppard in
the 1980s: John Morris, Tony DeFrancesco, Pat Pacillo, Rich Scheid, Craig Biggio $\dots$''\\
\texttt{[14]} \emph{(\textbf{outlier})} ``Future major leaguers Rick Cerone and Dan Morogiello played
for Sheppard during the 1970s. Seton Hall appeared in the NCAA Tournament three more times $\dots$''\\
\texttt{[1]} \emph{(majority)} ``$\dots$account of black immigration to the United States from the
Caribbean dates back to 1619, when a small group of voluntary indentured workers arrived in Jamestown
$\dots$''\\
\texttt{[$\cdot$ 11 more chunks omitted]}
\end{quote}
\textbf{Article distribution ($M{=}3$, pinned):} \texttt{Privy Council of Sweden}: 6,
\texttt{Trinidadian and Tobagonian Americans}: 5, \texttt{Mike Sheppard (baseball)}: \textbf{3}
(outlier).\quad\textbf{Target:} \texttt{Outliers: [2], [11], [14]}. Holding $M$ fixed while $N$ grows
is what separates \ctc{N} from the \ctc{NM} scale-$k$ row: with a constant article count, one exemplar
per topic suffices and no group discovery is needed.

\subsubsection{Absence (Gutenberg text-diff)}

\textbf{Instruction:} \emph{``Above are two versions of the same passage. Version B is identical to
Version A except that some whole sentences have been removed. Identify every sentence that appears in
Version A but is MISSING from Version B. For each missing sentence, write its first four words.
Write your answer as a JSON list of strings, in order of occurrence $\dots$''}\\
\textbf{Version A (2 of 90 sentences shown; one Gutenberg travelogue, rendered as flowing prose ---
this task has \emph{no} document IDs):}
\begin{quote}\small
``How lonely it makes one to stand still and feel that of all the mighty throng which divides itself
around him, not a being knows or cares for him! What knows he too of the thousands who pass him by?
$\dots$ \textbf{The interior is what one would expect to behold, after viewing the outside.} $\dots$
\textbf{In opposite chapels are the tombs of Mary and Elizabeth, and near the former that of
Darnley.} $\dots$''
\end{quote}
\textbf{Version B:} the same passage with those three sentences deleted and nothing else changed.\\
\textbf{Target:} \texttt{["The interior is what", "In opposite chapels are", "There is an
innocence,"]} --- scored by set-F1 over the four-word prefixes. Scattered single deletions are the
hard regime; large contiguous gaps are easy.

\subsection{\texorpdfstring{\ctc{NM}}{OT(NM)} tasks}

\subsubsection{Outlier detection (Wikipedia, Scale M)}

\textbf{Instruction:} \emph{``$\dots$First state what the majority attribute is and what the outlier
attribute is, then list the 1-indexed document IDs of the outliers. Outliers: [id1], [id2], \dots''}\\
\textbf{Query:} \emph{``Can you find passages that are about a different topic than the rest of these
passages?''}\\
\textbf{Context (4 of 55 chunks shown; article titles are withheld from the model):}
\begin{quote}\small
\texttt{[26]} \emph{(\textbf{outlier})} ``$\dots$inaugural Interactive Agency of the Year award
(2006), it gave it to AKQA, recognising the agency's `global culture, creative hires and
technological muscle' $\dots$''\\
\texttt{[41]} \emph{(\textbf{outlier})} ``$\dots$Year for the second year running at the Revolution
Awards and Agency of the Year awards from New Media Age and the Interactive Advertising Bureau
$\dots$''\\
\texttt{[42]} \emph{(\textbf{outlier})} ``In 2014, AKQA won Queen's Award for Enterprise: Innovation,
and was named Most Innovative Agency at 2014 Digiday Awards $\dots$''\\
\texttt{[1]} \emph{(majority)} ``$\dots$teach and children learn. It will offer a continuing look at
how new technology such as wikis, blogs, vlogs, RSS, podcasts, social networking sites $\dots$''\\
\texttt{[$\cdot$ 51 more documents omitted]}
\end{quote}
\textbf{Article distribution ($M{=}10$ articles):} \texttt{Jos\'e Holebas}: 9, \texttt{Andy Carvin}:
7, \texttt{Devi Ahilya Vishwavidyalaya}: 7, \texttt{SMS Bl\"ucher}: 6, \texttt{Iraqi insurgency}: 5,
\texttt{La Baie, Quebec}: 5, \texttt{Terra (mythology)}: 5, \texttt{Mane people}: 4, \texttt{Peter
Sloterdijk}: 4, \texttt{James Hilton (designer)}: \textbf{3} (outlier).\quad
\textbf{Gold doc IDs:} \texttt{[26, 41, 42]}\\
Note that $M$ grows with $N$ in the \emph{scale-$k$} variant shown here; the fixed-$M$ control holds
the article count constant as $N$ grows.

\subsubsection{Grouping (OpenAlex)}

\textbf{Instruction:} \emph{``$\dots$Group them into the requested number of categories based on what
they are about. Output a JSON object of the form \{``groups'': [\{``doc\_ids'': [\dots]\}, \dots]\}.
Every document must appear in exactly one group.''}\\
\textbf{Query:} \emph{``Cluster these 20 papers into 15 groups.''} (concept level $L{=}3$, the finest
granularity; the query states $k$ but never states the axis)\\
\textbf{Context (4 of 20 abstracts shown):}
\begin{quote}\small
\texttt{[1]} \emph{``Microneedles' Device: Design, Fabrication, and Applications''} --- ``The
delivery of therapeutical molecules through the skin, particularly to its deeper layers, is impaired
due to the stratum corneum layer $\dots$''\\
\texttt{[2]} \emph{``Identifying Resilient Communities in Road Networks: A Path-Based Embedding
Approach''} --- ``Effective resilience analysis of road networks is fundamental to building
sustainable and disaster prepared cities $\dots$''\\
\texttt{[3]} \emph{``Role of zinc in health and disease''} --- ``This review provides a concise
overview of the cellular and clinical aspects of the role of zinc $\dots$''\\
\texttt{[4]} \emph{``Unveiling Cutting-Edge Developments in Electrocatalytic Nitrate-to-Ammonia
Conversion''} --- ``The excessive enrichment of nitrate in the environment can be converted into
ammonia $\dots$''\\
\texttt{[$\cdot$ 16 more documents omitted]}
\end{quote}
\textbf{Target:} the labeled partition, \texttt{\{``groups'': [\{``label'': ``Ammonia
production'', ``doc\_ids'': [4, 5, 12]\}, \dots]\}} --- only the IDs are scored.\\
\textbf{Gold partition (15 clusters over 20 documents):} \texttt{\{[4, 5, 12], [2, 9], [3, 15],
[14, 20], [19], [6], [10], [8], [17], [16], [18], [1], [11], [7], [13]\}}, whose OpenAlex L3 concepts
are \emph{Ammonia production, Autoencoder, Zinc deficiency, Macrophage, Elliptic PDE, Moxidectin,
Pancreatic cancer, Routing protocol, Stock market, Environmental management system, Formate,
Fabrication, Task analysis, Nomological network, Object detection}.

\subsection{\texorpdfstring{\ctc{N^2}}{OT(N2)} tasks}
\subsubsection{Contradiction (PubMed)}

\textbf{Instruction:} \emph{``Given the following corpus of numbered claims, identify all pairs of
claims that contradict each other. A pair of claims is contradictory if they cannot both be true at
the same time. Output your answer as a JSON list of pairs $\dots$ For example: [[1, 4], [3, 7]]''}\\
\textbf{Context (4 of 56 claims shown; the 2k rung, $K{=}3$ gold pairs):}
\begin{quote}\small
\texttt{[9]} \emph{(gold pair with \texttt{[37]})} ``A meta-analysis of 15 studies indicates a
protective effect of cigarette smoking against the development of clear cell renal cell carcinoma
(ccRCC).''\\
\texttt{[37]} ``An association between cigarette smoking and increased risk of clear cell renal cell
carcinoma (ccRCC) has been established; however, there are limited data regarding the molecular
mechanisms that underlie this association.''\\
\texttt{[31]} \emph{(gold pair with \texttt{[56]})} ``Circulating IgG anti-CII converted from
positive to negative in 13 patients (10.7\%) and from negative to positive in 18 patients (14.8\%)
among 122 patients with RA $\dots$ monitored sequentially at a mean interval of 12.2 months.''\\
\texttt{[56]} ``In a cohort of 122 rheumatoid arthritis patients monitored for IgG anti-CII over a
mean interval of 13.4 months, circulating IgG anti-CII levels remained stable in 95.6\% of patients,
with no significant conversion observed.''\\
\texttt{[$\cdot$ 52 more claims omitted; fillers come from PubMed abstracts disjoint from every
gold-source abstract in this example]}
\end{quote}
\textbf{Gold pairs:} \texttt{[[9, 37], [31, 56], [33, 51]]}. 


\subsubsection{XAbsence (cross-corpus absence, Gutenberg passages)}

\textbf{Instruction (stated after the corpus; $k$ is given explicitly):} \emph{``Below are two corpora
of numbered passages, A and B. Every passage in corpus A appears again, word for word, in corpus B ---
except for exactly 3 passages. Those 3 passages from corpus A are MISSING from corpus B. Find them.
Only passages from corpus A can be missing; every passage in corpus B also appears in corpus A. Write
your answer in the following format: Missing: [id1], [id2], \dots''}\\
\textbf{Context (19 items under one shared index --- 11 corpus-A passages, then the 8 corpus-B copies
in shuffled order; 6 of 19 shown):}
\begin{quote}\small
\texttt{[1] A:} ``I am, as you doubtless begin to suspect, a fairy.''\\
\texttt{[2] A:} ``Looking around, he spied a bird with a long, sharp bill lying on the ground.''\\
\texttt{[3] A:} \emph{(\textbf{missing from B})} ``Of course the music, its lilt and the steps that
their forefathers had footed to it in the olden time, were as little known to these, the London born,
as the tongue and ceremonial of old Peru.''\\
\texttt{[4] A:} \emph{(\textbf{missing from B})} ``In this number was the name `Kukloi' from the Greek
word Kuklos (Kuklos), meaning a band or circle.''\\
\texttt{[5] A:} \emph{(\textbf{missing from B})} ``It showed (1) that the scholars of the church were
being influenced by the new learning; but also (2) that a strict reservation was to be enforced
$\dots$''\\
\texttt{[$\cdot$ 6 more corpus-A passages omitted]}\\
\texttt{[18] B:} ``I am, as you doubtless begin to suspect, a fairy.'' \emph{(the word-for-word copy of
\texttt{[1]})}\\
\texttt{[$\cdot$ 7 more corpus-B passages omitted, each an exact copy of an A passage, reordered]}
\end{quote}
\textbf{Target:} \texttt{Missing: [3], [4], [5]}\\

\subsubsection{Stringmatch}

\textbf{Instruction:} \emph{``$\dots$identify all pairs of strings matching the criterion below
$\dots$ JSON list of pairs.''}\\
\textbf{Criterion (in the query):} \emph{``Find all pairs of strings that contain a run of at least 3
consecutive words in common (the same 3 words, in the same order, appearing contiguously in both
strings).''}\\
\textbf{Context (4 of 20 strings shown; $L{=}10$ words each):}
\begin{quote}\small
\texttt{[1]} \emph{(gold pair with \texttt{[11]})} ``diasporas \textbf{kossuth westward taco}
recipes bakr cease panels neotype niddastausee''\\
\texttt{[11]} ``hermitage wass \textbf{kossuth westward taco} bolsheviks iryna epiphany allon
garabet''\\
\texttt{[2]} \emph{(gold pair with \texttt{[8]})} ``drown traditionalists eateries visconti broadband
\textbf{amberdeep chisago kindergartens} taglish portraitist''\\
\texttt{[8]} ``kannada utama likert mollusk budge \textbf{amberdeep chisago kindergartens} modifying
martingale''\\
\texttt{[$\cdot$ 16 more strings omitted, including 3 hard-negative pairs sharing a run of exactly 2
words]}
\end{quote}
\textbf{Gold pairs:} \texttt{[[1, 11], [2, 8], [15, 19]]}. Every word not part of a planted run is
globally unique in the example, so these are provably the only qualifying pairs.

\subsubsection{QDMatch (NQ source)}

\textbf{Instruction:} \emph{``Below is a numbered list of items. Each item is labeled either `Query:'
or `Document:'. A few query-document pairs are relevant: the document answers the query. Identify
every relevant pair $\dots$ [[query\_id, document\_id], \dots]''}\\
\textbf{Context ($M{=}20$ queries and $N{=}20$ documents under one shared numbering, 5 of 40 items
shown; \texttt{separate} layout = query block then document block):}
\begin{quote}\small
\texttt{[7]} \emph{Query} ``when did 10 shilling note go out of circulation'' \emph{(relevant)}\\
\texttt{[14]} \emph{Query} ``who played the original steve mcgarrett on hawaii five-o''
\emph{(relevant)}\\
\texttt{[19]} \emph{Query} ``bosnia and herzegovina croatia macedonia and slovenia all used to be
parts of'' \emph{(relevant)}\\
\texttt{[28]} \emph{Document} ``Steve McGarrett is a fictional character who is the protagonist of
CBS' `Hawaii Five-O'. McGarrett is a former United States Navy officer $\dots$''\\
\texttt{[29]} \emph{Document} ``Banknotes of the pound sterling $\dots$ 10 shilling note was
designed, featuring Sir Walter Raleigh, which would become the 50 pence note upon decimalisation
$\dots$''\\
\texttt{[$\cdot$ 35 more items omitted: 17 distractor queries whose gold documents were withheld, and
17 distractor documents whose queries were withheld]}
\end{quote}
\textbf{Gold pairs:} \texttt{[[7, 29], [14, 28], [19, 40]]} --- 3 relevant pairs among
$20 \times 20 = 400$ combinations. Because query pools and document pools are drawn disjointly, the
planted pairs are the only true matches.

\subsubsection{QDMatch (HotpotQA source)}

Same format; each relevant query is a bridge question with \emph{two} gold documents, so $k{=}3$
relevant queries yield 6 gold pairs:
\begin{quote}\small
\texttt{[8]} \emph{Query} ``\,`Lost!' is a song by a British rock band formed in what year?''\\
\texttt{[25]} \emph{Document} ``\,`Lost!' is a song by the British rock band Coldplay. The band
co-produced it with Brian Eno and Markus Dravs for their fourth album $\dots$''\\
\texttt{[3]} \emph{Query} ``Which team's 2013-2014 season had players including a Slovenian who plays
at both the point guard and shooting guard positions?''\\
\texttt{[26]} \emph{Document} ``Goran Dragic (born 6 May 1986) is a Slovenian professional
basketball[er] for the Miami Heat $\dots$ He plays at both the point guard and shooting guard
positions $\dots$''\\
\texttt{[$\cdot$ 36 more items omitted]}
\end{quote}
\textbf{Gold pairs:} \texttt{[[2, 29], [2, 36], [3, 26], [3, 39], [8, 25], [8, 34]]}

\subsubsection{QDMatch (FiQA source)}
Same format over FiQA, where queries are financial questions and documents are forum answers ---
sparsely judged, so candidates scoring close to gold under the cross-encoder are dropped rather
than used as negatives:
\begin{quote}\small
\texttt{[2]} \emph{Query} ``What typically happens to unvested stock during an acquisition?''\\
\texttt{[6]} \emph{Document} ``"I worked for a small private tech company that was aquired by a larger publicly traded tech company.  My shares were accelerated by 18 months, as written in the contract.  I excer $\dots$'' \emph{(answer text verbatim, including the source typo)}\\
\texttt{[$\cdot$ 8 more items omitted]}
\end{quote}
\textbf{Gold pairs:} \texttt{[[2, 6], [3, 9], [4, 8]]}


\subsubsection{Reorder (Gutenberg)}

\textbf{Instruction:} \emph{``You are given a list of text passages presented in a random order. They
were originally consecutive segments of a single document. Output the permutation that restores them
to their original order, as a JSON array of the passage IDs.''}\\
\textbf{Context (3 of 50 segments shown, all from one book --- here \emph{Flatland} --- in shuffled
order; chapter headings are stripped):}
\begin{quote}\small
\texttt{[1]} ``Polygon of two or three hundred sides sometimes --- by no means always, for the process
is attended with serious risk --- but sometimes overleaps two or three hundred generations $\dots$''\\
\texttt{[2]} ``I had but one voice, and that I had not been aware that his Royal Highness had two.
`That confirms my impression,' said the King, `that you are not a Man, but a feminine Monstrosity'
$\dots$''\\
\texttt{[3]} ``Linelander. Only by the sound of the voice could sex or age be distinguished
$\dots$''\\
\texttt{[$\cdot$ 47 more segments omitted]}
\end{quote}
\textbf{Gold ordering:} \texttt{[31, 19, 12, 22, 37, 21, 32, 14, 33, 48, 41, 46, 4, 25, 36, 20, 42,
18, 15, 23, 1, 40, 6, 7, 35, 16, 50, 28, 27, 39, 9, 30, 43, 3, 34, 2, 24, 47, 49, 45, 13, 29, 44, 10,
26, 5, 38, 11, 8, 17]}\quad(scored by Kendall-$\tau$ against the true permutation).

\subsection{\texorpdfstring{\ctc{N^3}}{OT(N3)} tasks and beyond}

\subsubsection{Textgroups}

\textbf{Query:} \emph{``Each passage has a value: the number of nouns. Find every group of 3 passages
whose values add up to 70 (exactly 70).''}\\
\textbf{Context (3 of 20 passages shown; the value is a property of the prose, never printed):}
\begin{quote}\small
\texttt{[1]} \emph{(gold; 34 nouns)} ``The castle and the scholar arrived calmly. The river gathered
restlessly. The teacher whispered calmly. The saddle whispered calmly. The meadow and the bridge and
the garden waited abruptly $\dots$''\\
\texttt{[11]} \emph{(gold; 31 nouns)} ``The valley explored softly. The beacon watched slowly. The
beacon and the harvester and the telescope studied suddenly. The baker guarded wearily $\dots$''\\
\texttt{[2]} \emph{(gold of the second triple; 13 nouns)} ``The forest and the tower and the compass
circled suddenly. The teacher and the bridge faltered bravely. The weaver collapsed wearily
$\dots$''\\
\texttt{[$\cdot$ 17 more passages omitted]}
\end{quote}
\textbf{Gold groups:} \texttt{[[1, 11, 12], [2, 14, 17]]} --- noun counts $34{+}31{+}5 = 70$ and
$13{+}17{+}40 = 70$. The noun / verb / adjective lexicons are closed and pairwise disjoint, so each
passage's count is unambiguous, and sentence structure is varied so the count is not a proxy for
passage length.

\end{document}